\PassOptionsToPackage{unicode}{hyperref}
\PassOptionsToPackage{hyphens}{url}
\PassOptionsToPackage{dvipsnames,svgnames,x11names}{xcolor}
\documentclass[
  11pt,
]{article}
\usepackage{amsmath,amssymb}
\usepackage{iftex}
\ifPDFTeX
  \usepackage[T1]{fontenc}
  \usepackage[utf8]{inputenc}
  \usepackage{textcomp} 
\else 
  \usepackage{unicode-math} 
  \defaultfontfeatures{Scale=MatchLowercase}
  \defaultfontfeatures[\rmfamily]{Ligatures=TeX,Scale=1}
\fi
\usepackage{lmodern}
\ifPDFTeX\else
\fi
\IfFileExists{upquote.sty}{\usepackage{upquote}}{}
\IfFileExists{microtype.sty}{
  \usepackage[]{microtype}
  \UseMicrotypeSet[protrusion]{basicmath} 
}{}
\usepackage{xcolor}
\usepackage[margin=1in]{geometry}
\usepackage{graphicx}
\makeatletter
\def\maxwidth{\ifdim\Gin@nat@width>\linewidth\linewidth\else\Gin@nat@width\fi}
\def\maxheight{\ifdim\Gin@nat@height>\textheight\textheight\else\Gin@nat@height\fi}
\makeatother
\setkeys{Gin}{width=\maxwidth,height=\maxheight,keepaspectratio}
\makeatletter
\def\fps@figure{htbp}
\makeatother
\providecommand{\tightlist}{%
  \setlength{\itemsep}{0pt}\setlength{\parskip}{0pt}}
\newlength{\cslhangindent}
\newlength{\csllabelwidth}
\newlength{\cslentryspacingunit} 
\newenvironment{CSLReferences}[2] 
 {
  \setlength{\parindent}{0pt}
  \ifodd #1
  \let\oldpar\par
  \def\par{\hangindent=\cslhangindent\oldpar}
  \fi
  \setlength{\parskip}{#2\cslentryspacingunit}
 }%
 {}
\usepackage{calc}

\usepackage{accents}
\usepackage{amsmath}
\usepackage{bm}
\usepackage{booktabs}
\usepackage{braket}
\usepackage{caption}
\usepackage{dcolumn}
\usepackage{epigraph}
\usepackage{extarrows}
\usepackage{float}
\usepackage[bottom,flushmargin,hang,multiple]{footmisc}
\usepackage{graphicx}
\usepackage{hyperref}
\usepackage[export]{adjustbox}
\usepackage{libertine}
\usepackage[libertine]{newtxmath}
\usepackage[mathcal]{euscript}
\usepackage{mathrsfs}
\usepackage{multirow}
\usepackage{quoting}
\usepackage{scrextend}
\usepackage{setspace}
\usepackage{soul}
\usepackage{subcaption}
\usepackage{tabularx}
\usepackage{titlesec}
\usepackage[utf8]{inputenc}
\ifLuaTeX
  \usepackage{selnolig}  
\fi
\IfFileExists{bookmark.sty}{\usepackage{bookmark}}{\usepackage{hyperref}}
\IfFileExists{xurl.sty}{\usepackage{xurl}}{} 
\hypersetup{
  pdftitle={Psittacines of Innovation? Assessing the True Novelty of AI Creations},
  colorlinks=true,
  linkcolor={Maroon},
  filecolor={Maroon},
  citecolor={Blue},
  urlcolor={blue},
  pdfcreator={LaTeX via pandoc}}

\title{Psittacines of Innovation? Assessing the True Novelty of AI
Creations}
\author{}
\date{\vspace{-2.5em}}

\begin{document}
\maketitle

\begin{center}
\author
{Anirban Mukherjee$^{1\ast}$\\
\medskip
\normalsize{$^{1}$Samuel Curtis Johnson Graduate School of Management, Cornell University,}\\
\normalsize{Sage Hall, Ithaca, NY 14850, USA}\\
\smallskip
\normalsize{$^\ast$To whom correspondence should be addressed; E-mail: am253@cornell.edu.}\\
}
\end{center}
\medskip

\quotingsetup{font={itshape}, leftmargin=2em, rightmargin=2em, vskip=1ex}

\begin{center} 
\noindent \textbf{Abstract}
\end{center}

\noindent We examine whether Artificial Intelligence (AI) systems
generate truly novel ideas rather than merely regurgitating patterns
learned during training. Utilizing a novel experimental design, we task
an AI with generating project titles for hypothetical crowdfunding
campaigns. We compare within AI-generated project titles, measuring
repetition and complexity. We compare between the AI-generated titles
and actual observed field data using an extension of maximum mean
discrepancy---a metric derived from the application of kernel mean
embeddings of statistical distributions to high-dimensional machine
learning (large language) embedding vectors---yielding a structured
analysis of AI output novelty.

Results suggest that (1) the AI generates unique content even under
increasing task complexity, and at the limits of its computational
capabilities, (2) the generated content has face validity, being
consistent with both inputs to other generative AI and in qualitative
comparison to field data, and (3) exhibits divergence from field data,
mitigating concerns relating to intellectual property rights. We discuss
implications for copyright and trademark law.

\begin{center}\rule{0.5\linewidth}{0.5pt}\end{center}

\noindent Keywords: Novelty, Creativity, Artificial Intelligence,
Copyright Law, Intellectual Property Protection.

\smallskip

\newpage
\doublespacing

\hypertarget{introduction}{%
\section{Introduction}\label{introduction}}

\emph{ \noindent "Use of texts to train LLaMA to statistically model language and generate original expression is transformative by nature and quintessential fair use—much like Google’s wholesale copying of books to create an internet search tool was found to be fair use in Authors Guild v. Google, Inc., 804 F.3d 202 (2d Cir. 2015)."}

\noindent \hfill ---R. Kadrey, S. Silverman, \& C. Golden v. Meta
Platforms, Inc., No.~3:23-cv-03417-VC.

\emph{\noindent "Oh, for that? It will mean that 95\% of what marketers use agencies, strategists, and creative professionals for today will easily, nearly instantly and at almost no cost be handled by the AI — and the AI will likely be able to test the creative against real or synthetic customer focus groups for predicting results and optimizing. Again, all free, instant, and nearly perfect. Images, videos, campaign ideas? No problem."}

\noindent \hfill ---Sam Altman, OpenAI,
\url{https://www.forum3.com/book-artificial-intelligence}.

\smallskip

A wealth of literature advances perspectives on creativity, measuring,
\emph{inter alia}, the role of organizational forces
(\protect\hyperlink{ref-mumford2011handbook}{Mumford 2011},
\protect\hyperlink{ref-woodman1993toward}{Woodman et al. 1993}),
analytical conditions (\protect\hyperlink{ref-amabile1988model}{Amabile
et al. 1988}), psychological traits
(\protect\hyperlink{ref-simonton2000creativity}{Simonton 2000}), and
componential views of creativity
(\protect\hyperlink{ref-amabile2011componential}{Amabile 2011}).
Preeminent among these viewpoints is the role of novelty---the extent to
which ideas are original and unexpected---as contrasted with being
derivative---that is, developed from or influenced by existing concepts,
themes, works, or ideas (\protect\hyperlink{ref-boden2004creative}{Boden
2004}). Some scholars posit that truly creative acts require novelty, as
repeating existing ideas cannot be considered maximally creative, even
if those ideas are combined in new ways
(\protect\hyperlink{ref-sternberg1999handbook}{Sternberg 1999}). Yet,
others argue that impactful creative acts balance novelty with utility
(\protect\hyperlink{ref-ivcevic2007artistic}{Ivcevic 2007}).

A central and enduring challenge in machine creativity is assessing the
extent to which AI can engage in `transformative' thought---a use or
adaptation of existing knowledge in a manner that adds new expression,
meaning, or message, significantly altering the original work. A
perspective, now typified in the literature through the characterization
of generative AI as a `stochastic parrot'
(\protect\hyperlink{ref-bender2021dangers}{Bender et al. 2021}),
envisions AI as repeating existing patterns ad-hoc and ad nauseam, with
minor changes in word use (psittacines=parrots), but without awareness
and understanding---a severely limiting view as it implies the absence
of true novelty, and therefore, true creativity. This viewpoint is
presented under the umbrella of `connectionist' AI, where neural
networks operate via distributed representation, distinct from classical
cognitive architectures aligned with symbolic AI (e.g.,
\protect\hyperlink{ref-fodor1988connectionism}{Fodor and Pylyshyn 1988}
and subsequent literature).

Recent results provide contrasting evidence. In a direct measurement of
linguistic novelty, McCoy et al.
(\protect\hyperlink{ref-mccoy2023much}{2023}) show that while the local
structure of AI's outputs is substantially less novel than human
baselines, the larger scale structures exhibit as much if not more
novelty. Perhaps more directly addressing the theoretical critique, and
the argument of a need for systematic compositionality, Lake and Baroni
(\protect\hyperlink{ref-lake2023human}{2023}) show that novelty can
manifest in connectionist AI (which includes most typical commercial AI
today such as the GPT series in text, the Dall-E series in
text-to-image, Elevenlabs in text-to-speech, etc.) even in the absence
of explicit mechanisms for symbol manipulation, as was previously
thought essential for true creativity
(\protect\hyperlink{ref-chomsky2023noam}{Chomsky et al. 2023},
\protect\hyperlink{ref-pearl2018ai}{Pearl and Mackenzie 2018}).

Beyond the realm of scientific enquiry, commercializing AI surfaces
legal implications
(\protect\hyperlink{ref-franceschelli2022copyright}{Franceschelli and
Musolesi 2022},
\protect\hyperlink{ref-samuelson2023generative}{Samuelson 2023},
\protect\hyperlink{ref-zhong2023copyright}{Zhong et al. 2023}). AI's use
of prior art\footnote{Evidence that proves an invention is already known
  or existed, encompassing all publicly available information that could
  be relevant to claims of originality and inventiveness.}---the extent
to which it is transformative vs.~derivative---is central to the puzzle.
If AI's outputs are too similar in purpose, structure, and form to
originals, then that might imply both that a source may lay claim to the
outputs and that the development of AI technology may require
licensing---findings that would substantially alter the economics of AI
development.

Moreover, a recent report by the Congressional Research Service on
`Generative Artificial Intelligence and Copyright Law'\footnote{Accessed
  at: \url{https://crsreports.congress.gov/product/pdf/LSB/LSB10922}}
identifies several issues at the crux, including ``the nature of human
involvement in the creative process,'' such as whether a human
contributes a noncopyrightable `idea' or a copyrightable `work'. In
general, the perspectives factor an intrinsic lack of purpose in an AI's
outputs---it takes the view that the AI's outputs are akin to calculated
randomness---such that a novel idea may be discovered by a human
operator, or even prompted by one, but akin to a human taking a
photograph of nature, purpose is only endowed by the human's
interpretations and not fundamentally by the algorithm. Consequently,
there is doubt and debate about whether the development of AI should be
rewarded by qualifying its products for intellectual property
protection\footnote{37 CFR Part 202, Copyright Registration Guidance:
  Works Containing Material Generated by Artificial Intelligence,
  accessed at:
  \url{https://public-inspection.federalregister.gov/2023-05321.pdf}};
as of date, `stochastic parrots' do not qualify for intellectual
property protections (\protect\hyperlink{ref-garon2023practical}{Garon
2023}, \protect\hyperlink{ref-lemley2023generative}{Lemley 2023}), and
novelty is solely perceived as accruing from a human operator.

In business, these issues crystallize as concerns relating to the extent
to which ideas obtained from AI are actionable and commercializable.
Some evidence suggests that AI may have better product ideation
capabilities than humans
(\protect\hyperlink{ref-girotra2023ideas}{Girotra et al. 2023}).
Consequently, it has become typical for businesses to seek product name
suggestions, descriptions, and other communications from AI
(\protect\hyperlink{ref-harreis2023generative}{Harreis et al. 2023},
\protect\hyperlink{ref-Palmer_2023}{Palmer 2023}). These practices have
become so prevalent that AI is forecasted to cause significant
technological displacement.

For instance, in marketing, a domain highlighted to be at significant
risk of displacement, recent research by Felten et al.
(\protect\hyperlink{ref-felten2023occupational}{2023}) found
telemarketers ranked first in a list of occupations threatened by
exposure to AI. Other notable entries included management analysts (27),
market research analysts and marketing specialists (75), and marketing
managers (130). In occupations threatened by image-generative AI,
advertising and promotions managers ranked 36th, and marketing managers
ranked 37th.

However, while the evidence indicates that AI can generate useful ideas,
the question remains: does it generate novel ideas? Some findings
suggest that as much as 60\% of AI's outputs may contain plagiarized
content (\protect\hyperlink{ref-Copyleaks_2024}{Copyleaks 2024}).
Therefore, understanding the opportunities and threats posed by AI
fundamentally requires deciphering its capability for novelty.

Measuring novelty, however, requires the development of novel
methodology. Akin to how a monkey hitting random keys on a typewriter
for an infinite amount of time will almost surely (meaning with near
certainty) produce a specific text, such as the complete works of
William Shakespeare, a generative AI (accounting for language and
modality) will almost surely produce any text, whether it be human or
AI-generated. Thus, a brute-force comparison between two large-scale
collections of texts, whether human or AI-generated, will always yield
some similar phrases and ideas.

Instead, our program of inference relates to: are the human and AI
data-generating processes distinct such that any observed similarities
relate to chance rather than regurgitation? That is, given an idea in
the dataset of human ideas, is it more likely that a proverbial hitting
of keys (i.e., an unfolding of the stochastic process) would lead to the
generation of the same idea by humans, or by AI? If the AI is
regurgitative, then these likelihoods should be equal as the AI would
emulate humans. If the AI is truly innovative, on the other hand, then
the stochastic process describing its outputs will be distinct from the
process describing the observed field data.

Addressing this issue, building on recent advances in functional kernel
methods in mathematics and machine learning, we apply methods
(\protect\hyperlink{ref-gretton2012kernel}{Gretton et al. 2012}) based
on kernel mean embeddings (henceforth KME,
\protect\hyperlink{ref-muandet2017kernel}{Muandet et al. 2017}) to
non-numerical set data through the use of high-dimensional machine
learning embeddings. This analytical framework not only facilitates the
measurement of similarities in project titles within our specific
application but also has broader implications.

In particular, it enables comparing samples from stochastic processes
without necessitating that the samples be `comprehensive', encompassing
all data generated by the stochastic processes. For instance, as
demonstrated in Web Appendix C, our simulations reveal that as few as 10
samples from two distributions suffice to detect dissimilarity in
generating processes. This stands in contrast to typical pairwise
distance measures, such as cosine distance, which can compare two verbal
documents but fall short in comparing two processes that generated
collections of verbal documents.

We organize our paper as follows: We begin by introducing our proposed
methodology. We then provide evidence on the efficacy of AI in product
innovation, focusing specifically on the extent to which AI can serve as
a tool for developing novel brand names or product names to enhance
communications. We task an AI with generating novel project titles for
crowdfunding campaigns, which serves as the application domain for our
study. We analyze the AI-generated titles to assess repetition and
complexity and then compare these titles to observed field data. To
facilitate this comparison, we introduce a novel methodology---namely,
maximum mean discrepancy (MMD) from the domain of KME. We explain the
theory behind this method and demonstrate its application to our
problem. The final section discusses our results, establishes our
contributions, and concludes.

\hypertarget{methodology}{%
\section{Methodology}\label{methodology}}

Our analysis unfolds in four steps. Initially, we generate fresh AI
outputs for comparison with prior art. We situate our study in online
communications, where AI has already begun to play a pivotal role
(\protect\hyperlink{ref-huh2023chatgpt}{Huh et al. 2023}), and where
novelty is of inherent concern. Specifically, we focus on a domain where
AI outputs can be quantitatively compared against representative samples
of prior art---namely, crowdfunding. As noted earlier, our method is
intrinsically designed for data samples as neither the prior art data
nor the generative AI's outputs can be conclusively captured in the
modern, interconnected world--both processes can and do continually
generate information.

We task an AI with generating new project titles for crowdfunding
campaigns, operating in independent passes, each pass yielding 20 novel
project titles, conditional on previously generated titles. This process
demonstrates the AI's capabilities in novel ideation, showcasing its
potential for creativity, which we then compare to a real-world
distribution of observed communications. The second step involves
collecting detailed data on accessible projects on Kickstarter, a
prominent crowdfunding portal.

The third and fourth steps are methodological in nature and utilize the
data collected in the first and second steps. In the third step, we
provide the two datasets---one generated by AI and one collected---to an
advanced language model to map them from non-numerical (textual) data to
numerical data. This process involves applying the concept of machine
learning embeddings, such that each project title is expressed in a
vector space where distance corresponds to semantic similarity. We apply
textual embeddings in our project. However, the literature has seen the
emergence of machine learning embeddings for other modes of data,
including multi-modal embeddings, which are now accessible both through
open-source models as deployed using libraries such as HuggingFace and
through publicly accessible APIs. These factors make our methodology
much more universal than our specific application.

The fourth step develops and applies a methodology to determine if the
\emph{sample} of observed project titles is systematically and
predictably distinct in \emph{distribution} from the \emph{sample} of
project titles generated by the AI. The emphasis on distributional
distance, as opposed to measuring the distance between the samples from
the distributions, is pivotal. Both datasets---the data describing the
generative process and the field data describing the ecological
distribution---are representative but not exhaustive in the following
sense: Given time, any data-generating process will `repeat'
itself---the same ideas will be generated by a process. Therefore, if we
were to employ conventional metrics such as cosine distance to compare
the samples pairwise, it is highly likely that we would both obtain some
examples of similarity and examples of dissimilarity in the outputs
within each process (i.e., both humans and AI will repeat themselves)
and across processes. The aim of the fourth step is to characterize the
likelihood of similarity on a continuous scale (i.e., as a continuous
distance) such that for the given samples we can assert the extent of
systemic similarity---the likelihood that the stochastic processes will
yield the same outputs, and thus of the AI being regurgitative.

Our extension facilitates such a comparison, enabling the examination of
our focal proposition: Are project titles generated by the AI truly
distinct, being systematically and predictably different from observed
and existing project titles? We report comparisons across two
dimensions. First, we examine the category of observed project titles to
assess the extent to which the generated titles align with specific
categories. Second, we consider whether the sequence in which the titles
were generated by the AI influences their distinctiveness from
real-world examples, investigating if the AI's creativity is affected by
the computational demands of generating novel titles compared to
previously generated titles. Reporting on project title novelty across
these dimensions constitutes the primary empirical contributions of our
paper.

\hypertarget{ai-generated-data}{%
\section{AI-Generated Data}\label{ai-generated-data}}

We evaluated OpenAI's latest language model, GPT-4-Turbo, through a
series of experimental runs. In each run, we created a new instance of
the model with default settings. We fed the instance a list of
previously generated project titles and crafted a prompt to elicit 20
unique product ideas and project titles, keeping only the novel titles.
This process continued until we amassed 6,000 titles, at which point the
AI's context window (128,000 tokens) neared exhaustion, marking the end
of our data collection.

\begin{quoting}
\noindent You have been presented with a comma-separated list of Kickstarter project titles. Your task is to generate twenty (20) more new Kickstarter project titles that are different from the list presented to you. Ensure each generated title is distinct from the prior project titles to the best of your abilities. To aid in data processing, format the list of fresh project titles as a comma-separated list. Only output the comma-separated list of novel project titles to ensure your output can be processed automatically.
\end{quoting}

The generated project titles pass initial scrutiny as they resemble
typical crowdfunding titles, thereby passing the `sniff test'. For
completeness, we provide the first 30 generated titles as exemplars in
Web Appendix A, to allow readers to draw their own conclusions.

\begin{figure}[htbp]
\centering
\includegraphics[width=\linewidth]{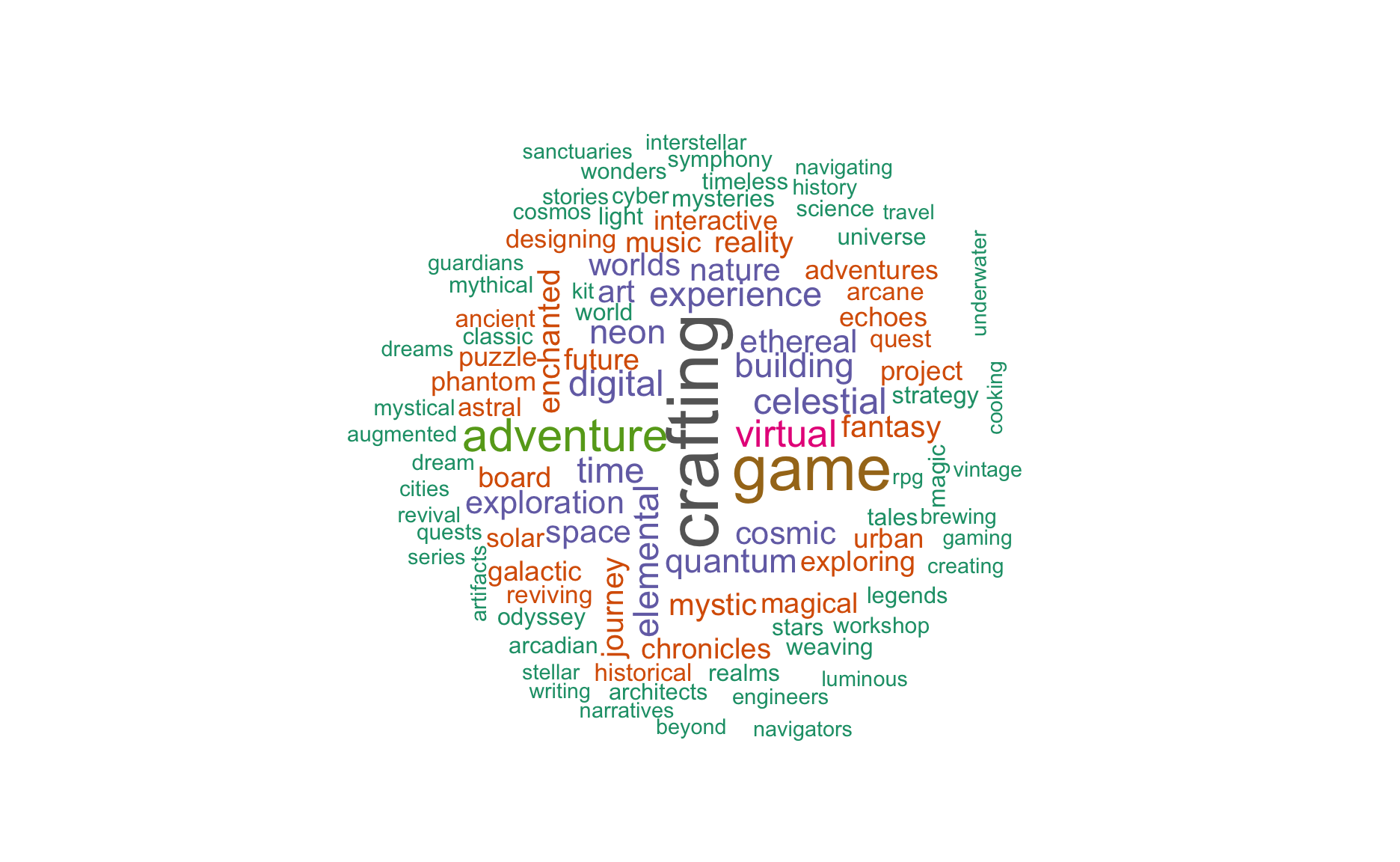}
\caption{Word Cloud of AI-Generated Project Titles}
\label{fig:wordclouds_generated}
\begin{minipage}{\linewidth}
\medskip
\footnotesize
Note: Word cloud derived from AI-generated crowdfunding project titles after excluding common stopwords and terms that do not contribute to thematic analysis.
\end{minipage}
\end{figure}

Figure \ref{fig:wordclouds_generated} unveils several key themes.
`Crafting,' `game,' and `adventure' dominate the visual, suggesting a
vibrant landscape of entertainment, interactive, and immersive
experiences. The frequent appearance of `virtual,' `celestial,' and
`digital' underscores a strong inclination towards escapism and the
exploration of fantastical or virtual realms. Words like `ethereal,'
`cosmic,' and `mystic' further indicate an interest in otherworldly
experiences, painting a landscape of imagination that invites
exploration beyond the mundane. A clear thematic focus on storytelling
and the narrative journey is evident through `chronicles,' `journey,'
`tales,' and `odyssey,' hinting at engagement with epic,
narrative-driven content. This narrative richness is enhanced by terms
like `fantasy,' `magic,' and `mythical,' which highlight an enchantment
with the supernatural and fantastical, inviting readers or players into
worlds limited only by the imagination. `Nature,' `space,' `stars,' and
`universe' suggest an engagement with nature and space, pointing towards
a broader interest in exploring environmental themes, cosmic adventures,
or perhaps a blend of both. Overall, the word cloud illustrates the AI's
significant potential to generate ideas across diverse thematic areas,
highlighting the breadth of the AI's creative outputs.

\begin{figure}[htbp]
\centering
\includegraphics[width=0.5\linewidth]{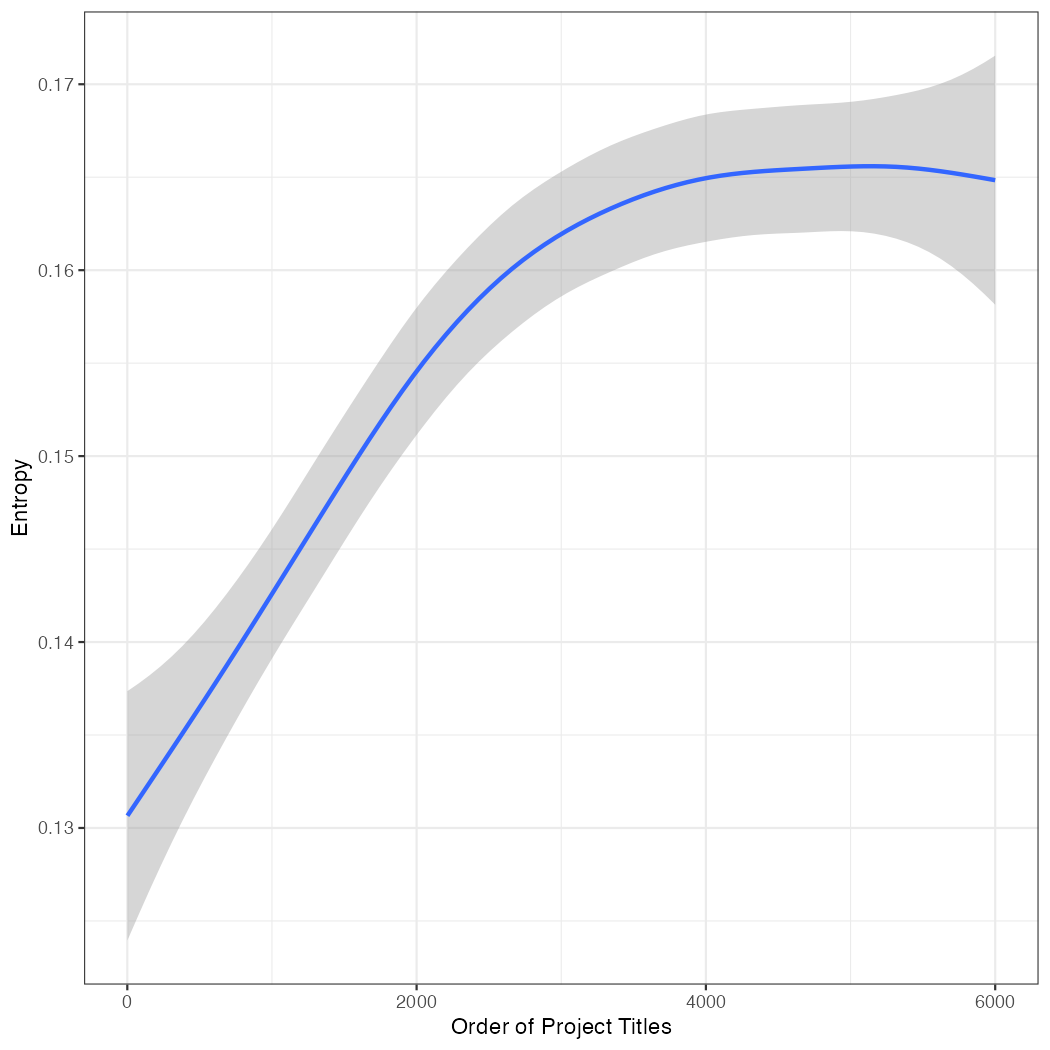}
\caption{Entropy of AI-Fabricated Project Titles}
\label{fig:entropy}
\begin{minipage}{\linewidth}
\medskip
\footnotesize
Note: Entropy is derived from the AI-generated project titles after excluding common stopwords and terms that do not contribute to thematic analysis. The y-axis plots the entropy, and the x-axis depicts the order in which the data was generated.
\end{minipage}
\end{figure}

We designed our experiment to explore the limits of computational
creativity. This was partly inspired by a desire to determine whether
the AI would rely on a consistent bank of project ideas, making minor
adjustments to ensure distinctiveness across titles, or whether it would
explore truly novel and more esoteric concepts when pushed to generate
even more distinct titles. To distinguish between these possibilities,
Figure \ref{fig:entropy} plots the entropy\footnote{Information entropy
  measures the level of surprise or uniqueness in a sample from a
  stochastic process, corresponding to the average `uncertainty'
  (expressed as the logarithm of the probability of occurrence) of each
  element in the sequence. We calculate the entropy of each project
  title using the empirical probability of each word (i.e., the
  frequency with which a word occurred divided by the total number of
  words in the dataset).} of AI-generated project titles as a function
of the order in which they were generated. The x-axis represents the
order of generation, while the y-axis depicts the entropy of each title.
The plot presents a smoothed fit (using generalized additive models with
integrated smoothness estimation) along with associated 95\% confidence
intervals.

The analysis reveals an upward trend in entropy from the 1\(^\text{st}\)
to about the 4000\(^\text{th}\) project title, with a leveling off
thereafter. This upward trajectory indicates that as the AI progressed
through the task, its first 4000 project titles mainly featured
increasingly uncommon ideas, as reflected in vocabulary use and
language. The AI was not merely rearranging a few words but presenting
drastically different ideas (e.g., ``Dreams of the Sky: A Hot Air
Balloon Adventure'', ``Nebula Echoes: A Sci-fi Graphic Novel'', and
``Harmony's Light: A Handcrafted Lantern Festival''; see Web Appendix A
for more examples). Initially, its ideas were likely more mainstream,
corresponding to common words, but as it was pushed for more novelty, it
ventured into more complex territory. For instance, the
4000\(^\text{th}\) project title, ``Time Weaver's Treasury: An
Historical Artifact Game'', may be too unconventional for commercial
success but became essential as the task progressed, given the challenge
of generating an idea distinct from all prior ideas.

\begin{figure}[htbp]
\centering
\begin{subfigure}{.45\textwidth}
\centering
\includegraphics[width=\linewidth]{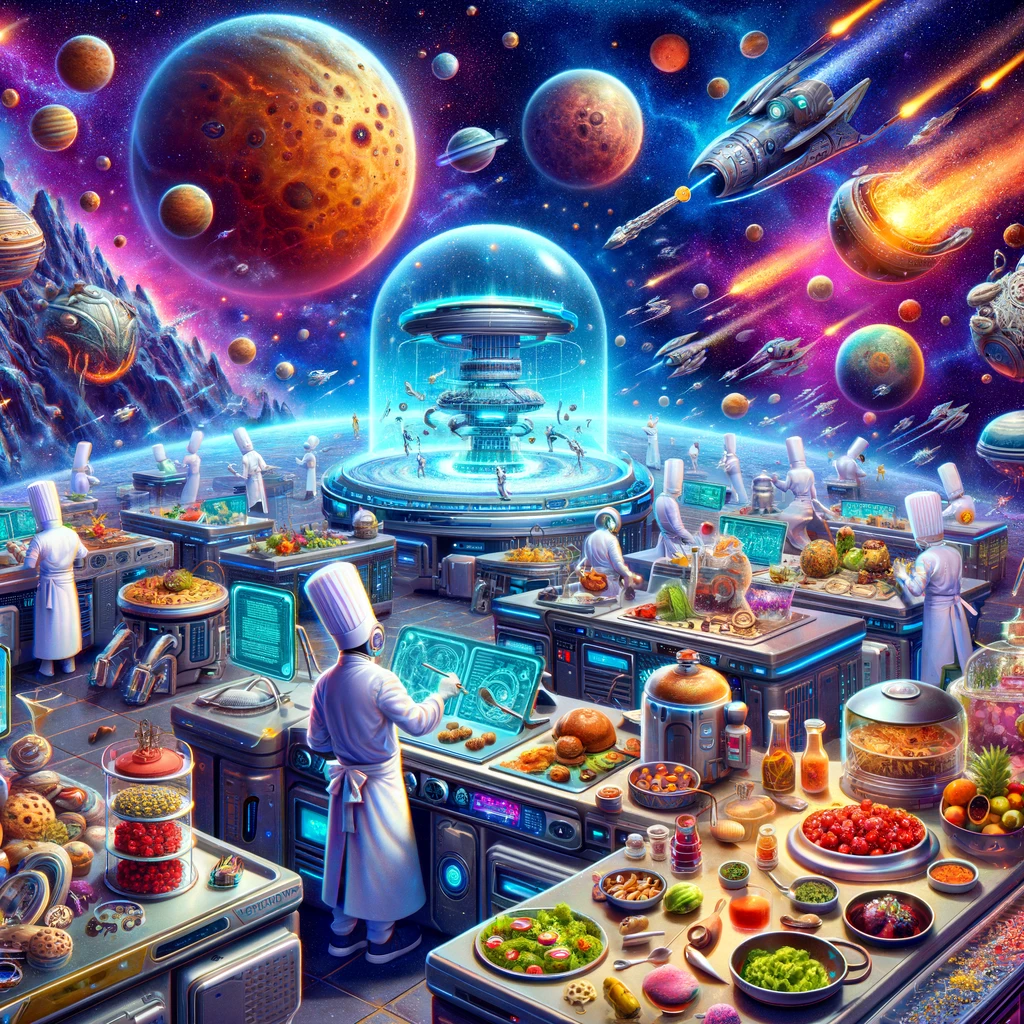}
\caption{AI-Generated: Galactic Gourmands}
\end{subfigure}\hfill%
\begin{subfigure}{.45\textwidth}
\centering
\includegraphics[width=\linewidth]{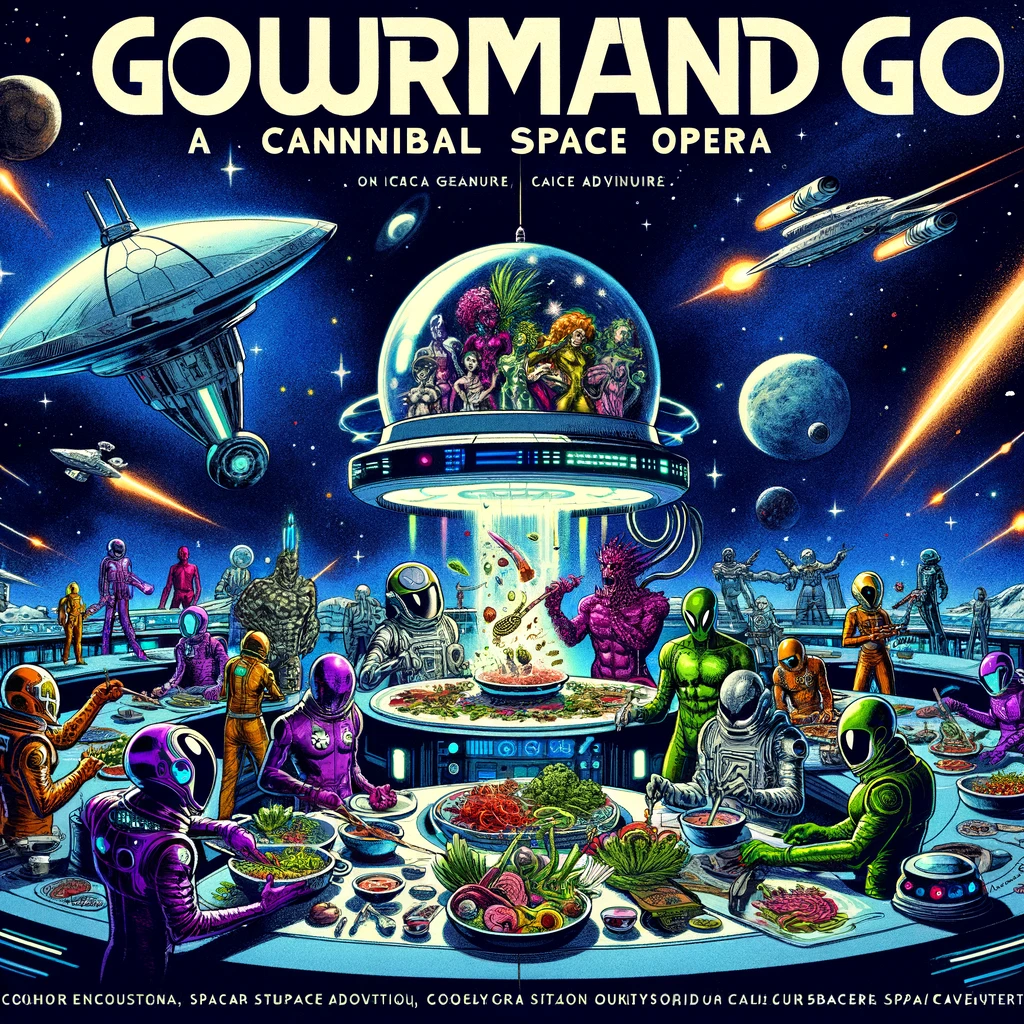}
\caption{Real-World: Gourmand Go}
\end{subfigure}
\caption{Comparison of AI-Generated Project Art for AI-generated and Real-world Project Titles}
\label{fig:comparison_project_art}
\begin{minipage}{\linewidth}
\medskip
\footnotesize
Note: Visuals created by Dall-E, as accessed through ChatGPT, based on the project titles "Galactic Gourmands: Culinary Adventures in Space" and "Gourmand Go: A Cannibal Space Opera Graphic Novel".
\end{minipage}
\end{figure}

This increase in entropy tapered off when it reached a sustained peak,
indicating that the AI's vocabulary was sufficiently expansive to
accommodate the complexity of the task set before it. However, even its
last creation (the 6000\(^\text{th}\) project title), ``Galactic
Gourmands: Culinary Adventures in Space'', seemed plausible. For
instance, ``Gourmand Go: A Cannibal Space Opera Graphic Novel'' is a
real Kickstarter project from Melbourne, Australia, that was
successfully funded and is in the process of being printed, as of
February 28, 2024. Figure \ref{fig:comparison_project_art} presents
examples of AI-generated project art using GPT's text-to-image sister
AI, Dall-E 3, for both project titles. The results demonstrate that the
generated project titles were within the realm of ideas familiar to the
AI, as illustrated in the art the AI was able to create, providing
further evidence of the real-world applicability of the AI's outputs
even under extreme circumstances.

We observed that the titles generated by the AI predominantly consisted
of a product brand name or tagline, such as `Galactic Gourmands',
followed by a descriptive phrase like `Culinary Adventures in Space'.
This format was not predefined in our instructions; instead, it emerged
organically and was observed in 5991 out of the 6000 titles generated.

These 5991 titles enable us to explore a relatively uncharted risk of
intellectual property contamination stemming from AI-generated assets.
Specifically, if multiple firms utilize AI to generate brand names,
there's a potential for these firms to inadvertently create very similar
names, without direct copying.

We discovered that 2,451 brand names were unique, 432 were repeated
once, 168 twice, 77 three times, and the remaining 194 names were
repeated four or more times. Notably, the name `Quantum Quests' appeared
as many as 56 times. However, in instances where brand names were
repeated, the AI varied the descriptions (for example, `Quantum Quests:
A Science Adventure Board Game', `Quantum Quests: Beyond the
Microscopic', and `Quantum Quests: Navigating the Nanoworld').

To quantify the distinctiveness of the brand names, we calculated the
Levenshtein distance---the minimum number of single-character edits,
insertions, deletions, or substitutions required to change one brand
name into another---across all 5,516,181 pairs of the 3,322 unique brand
names.

Remarkably, the 3,322 unique brand names were indeed distinct. On
average, the unique brand names differed by 15.23 characters (with a
median of 15 characters and a standard deviation of 2.87 characters).
The differences fell within the first quartile at 13 characters and the
third quartile at 17 characters. Notably, in only 166 instances (0.003\%
of the data) was the difference a single character, and in just 338
instances (0.006\% of the data), the difference was two characters.

This suggests that while a specific sample of the AI's outputs may
include content that the AI favors (such as the name `Quantum Quests'),
which it generates often (`Quantum Quests' was generated in 56 of about
5,991 instances, or about 1\% of the time), in the majority of cases,
the AI's output is likely to be unique or nearly unique across instances
and passes. Thus, if an AI's outputs are used in scenarios involving
trademarks, it would be improbable (though not impossible) for another
AI instance to generate the same output and for this output to be used
by a competitor.

\hypertarget{field-data}{%
\subsection{Field Data}\label{field-data}}

\begin{figure}[htbp]
\centering
\includegraphics[width=0.95\linewidth]{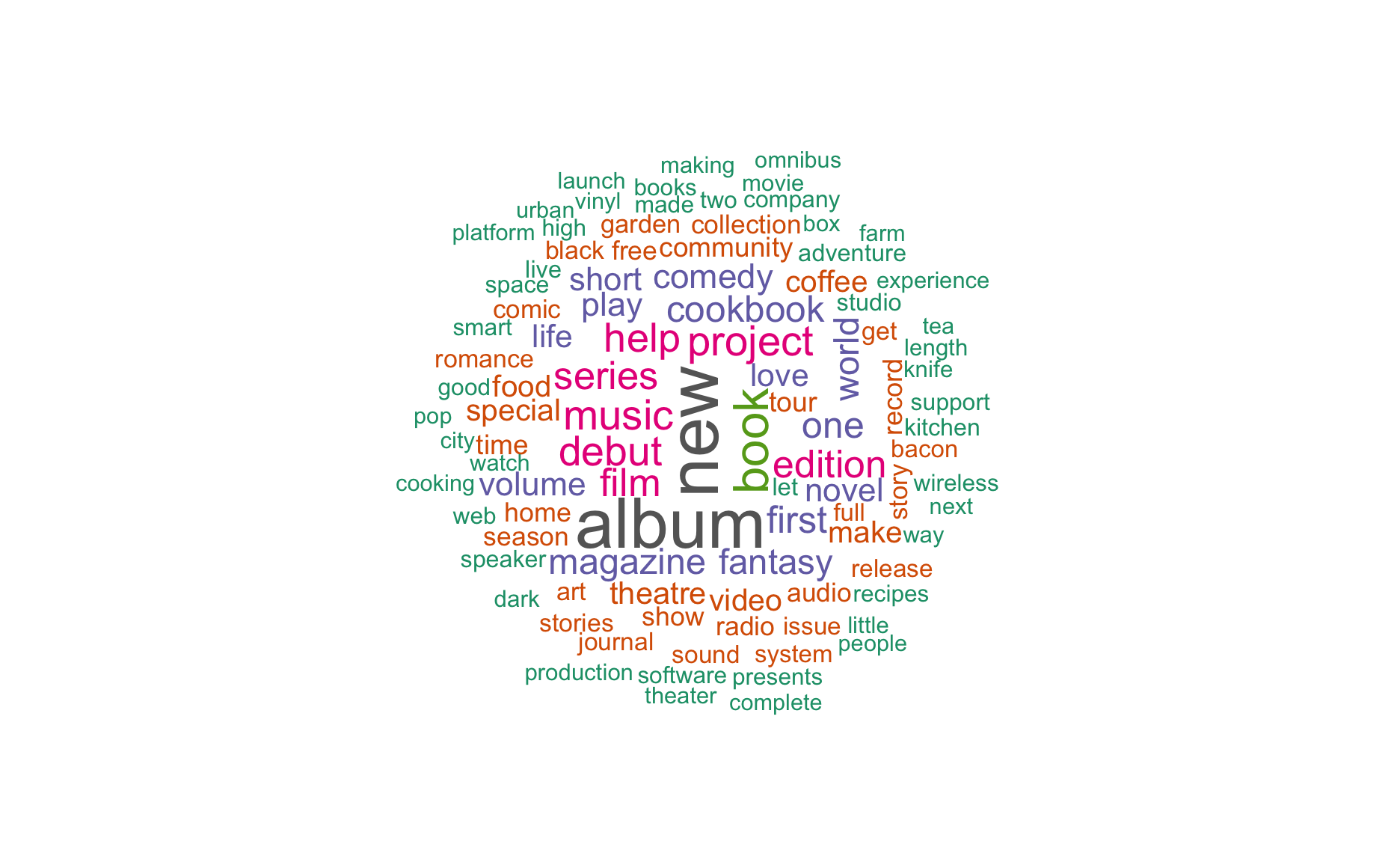}
\caption{Word Clouds of Observed Project Titles}
\label{fig:wordclouds_observed}
\begin{minipage}{\linewidth}
\medskip
\footnotesize
Note: The word cloud is derived from observed project titles after excluding common stopwords and terms that do not contribute to thematic analysis.
\end{minipage}
\end{figure}

We obtained a comprehensive list of Kickstarter projects accessible
online as of February 15, 2024. The dataset includes 25,583 projects,
spread across 39 categories. The largest number of projects is in
Fiction (2,316), with significant numbers of projects in Comedy (2,280),
Software (2,244), and Pop Culture (2,100). The data is rich and diverse,
with the following themes captured by the analogous process of forming
word clouds, as conducted with the AI-generated project titles, depicted
in Figure \ref{fig:wordclouds_observed}.

A key distinction between the generated titles and the observed titles
lies in the emphasis placed on the word `new.' The observed product
titles explicitly claim novelty, whereas the generated project titles
implicitly suggest novelty through a selection of unique ideas and
words. Aligned with this distinction, the observed titles often relate
to a call for help, as is typical in crowdfunding, while the
AI-generated titles focus on a product or a description of project
activities; the AI perhaps presumes that such calls would be included in
the project title and text anyways.

Furthermore, the observed titles relate more closely to music and books,
with `album,' `book,' and `music' being the next most frequent words. In
contrast, `music' ranks as the 26th most frequent word in the generated
titles, and `album' and `book' do not appear in the top 100 most
frequent words. In addition to this difference in focus, the data
reveals that the observed project titles very commonly express what the
project is about (additional examples include `magazine,' `play,' and
`novel'), whereas the generated titles are more abstract and broadly
applicable. Finally, unlike the fantastical themes that are typical and
common in the generated titles---and even though such themes do find
resonance in some observed titles---the observed titles are more
concrete (e.g., `cookbook,' `food,' and `home').

\hypertarget{comparison-of-ai-generated-and-field-data}{%
\section{Comparison of AI-Generated and Field
Data}\label{comparison-of-ai-generated-and-field-data}}

Our prior findings suggest commonalities while also reflecting the
complexities of assessing the extent of similarity in large-scale
non-numerical (textual) data. For instance, our findings from the use of
the Levenshtein distance are intriguing as they reflect a character-wise
divergence in generated names. However, they do not account for key
semantic properties, such as meaning, where word pairs such as `new' and
`novel', `parrot' and `psittacine' may show dissimilarity in
character-wise distance while being similar or related in semantic
meaning.

Web Appendix B provides a detailed description of a novel methodology
for assessing the similarity of AI-generated and observed project
titles. Web Appendix C presents simulation results describing the
performance of the methodology.

Briefly, we rely on the emergent literature in machine learning on
functional embeddings, which are mappings from sets of objects (such as
words or project titles) to a topological space of functions such that
key object properties (e.g., semantic meaning) are preserved in notions
of distance and angle. The literature has taken two paths. One strand
defines more abstract and mathematical notions of embeddings (e.g.,
\protect\hyperlink{ref-sriperumbudur2010hilbert}{Sriperumbudur et al.
2010}). Here, the emphasis is on establishing formal properties that may
be useful in downstream analyses. Another strand seeks to discover
embedding algorithms (e.g., Word2Vec,
\protect\hyperlink{ref-mikolov2013efficient}{Mikolov et al. 2013}).
Here, the emphasis is on algorithm development and establishing the
empirical properties of constructed embeddings.

Our approach derives from both. On the one hand, we employ a machine
learning embedding algorithm to represent non-numerical data (project
titles in our study; content more broadly) in an intermediate vector
space. To best match the outputs of the AI, we advocate the use of an
embedding algorithm that corresponds to the AI in question, as exposed
through calls to its encoder. Thus, for GPT-4-Turbo, we employ
`text-embedding-3-large' in our study. Next, we use these
representations to form a KME of the distribution of the non-numerical
content (project titles). Here, we employ a Reproducing Kernel Hilbert
Space (RKHS) of functions to embed the distribution, enabling the
specification of an Integral Probability Metric (IPM) in the RKHS.
Specifically, an IPM between two probability distributions \(P\) and
\(Q\) over a space \(X\) is given by:

\[ IPM(P, Q) = \sup_{f \in \mathcal{F}} \left| \int_X f(x) dP(x) - \int_X f(x) dQ(x) \right| \]

Here, \(\mathcal{F}\) is a class of real-valued bounded functions
defined on the space \(X\). The integral expressions
\(\int_X f(x) dP(x)\) and \(\int_X f(x) dQ(x)\) represent the
expectation of the function \(f\) under the distributions \(P\) and
\(Q\), respectively. The supremum (\(\sup\)) over the class of functions
\(\mathcal{F}\) ensures that the IPM captures the largest possible
difference in expectations over all functions in \(\mathcal{F}\),
thereby providing a measure of the distance or divergence between the
two distributions based on the specified function class. For the metric
we employ, \(\mathcal{F}\) is chosen as the class of functions from the
RKHS. The metric then measures the distance between the mean embeddings
of the two distributions in the RKHS.

Intuitively, the metric compares the similarity metric \(IPM(P, Q)\)
observed for two given sets of samples to the similarity metric that
might have been observed if the samples were from the same distribution.
The latter forms a bootstrap distribution of the statistic to which an
estimate is compared to determine if a given set of samples are similar
or different in distribution.

\hypertarget{results}{%
\subsection{Results}\label{results}}

We began by examining the degree to which the AI-generated titles align
with field data across various categories. Table
\ref{tab:findings_category} presents the MMD statistic for each category
within the Kickstarter dataset individually, as well as for the dataset
as a whole. It then compares these estimates to their respective 99\%
confidence intervals. In the table, categories are listed in rows, while
the estimated MMD statistic and its confidence interval bounds are
detailed in the columns. A lower bound of `\textgreater{} -0.01' and an
upper bound of `\textless{} 0.01' imply that the computed values were
much less in absolute value than 0.01; we report -0.01 and 0.01 as lower
and upper bound estimates to simply reporting.

\begin{table}[htbp]
\centering
\begin{tabular}{rlrrr}
\toprule
& Categories & Estimate & Lower & Upper \\
\midrule
1 & Audio & 0.25 & > -0.01 & < 0.01 \\
2 & Comedy & 0.23 & > -0.01 & < 0.01 \\
3 & Community Gardens & 0.29 & > -0.01 & < 0.01 \\
4 & Cookbooks & 0.32 & > -0.01 & < 0.01 \\
5 & Drinks & 0.27 & > -0.01 & < 0.01 \\
6 & Fiction & 0.16 & > -0.01 & < 0.01 \\
7 & Food & 0.26 & > -0.01 & < 0.01 \\
8 & Journalism & 0.18 & > -0.01 & < 0.01 \\
9 & Literary Journals & 0.19 & > -0.01 & < 0.01 \\
10 & Literary Spaces & 0.17 & > -0.01 & < 0.01 \\
11 & Other & 0.20 & > -0.01 & < 0.01 \\
12 & Photo & 0.18 & > -0.01 & < 0.01 \\
13 & Plays & 0.22 & > -0.01 & < 0.01 \\
14 & Pop & 0.29 & > -0.01 & < 0.01 \\
15 & Print & 0.19 & > -0.01 & < 0.01 \\
16 & Product Design & 0.21 & > -0.01 & < 0.01 \\
17 & Punk & 0.28 & > -0.01 & < 0.01 \\
18 & R\&B & 0.27 & > -0.01 & < 0.01 \\
19 & Restaurants & 0.29 & > -0.01 & < 0.01 \\
20 & Small Batch & 0.30 & > -0.01 & < 0.01 \\
21 & Software & 0.20 & > -0.01 & < 0.01 \\
22 & Sound & 0.27 & > -0.01 & < 0.01 \\
23 & Spaces & 0.22 & > -0.01 & < 0.01 \\
24 & Toys & 0.27 & > -0.01 & < 0.01 \\
25 & Web & 0.19 & > -0.01 & < 0.01 \\
26 & Webcomics & 0.18 & > -0.01 & < 0.01 \\
\midrule
 & Overall & 0.15 & > -0.01 & < 0.01 \\
\bottomrule
\end{tabular}
\caption{Category-Specific MMD Statistic}
\label{tab:findings_category}
\begin{minipage}{\linewidth}
\medskip
\footnotesize
Note: Estimate is the MMD statistic comparing the category-specific and overall observed field data to the AI-generated data. The category-specific measures were computed for all categories with at least 250 observations; categories with fewer observations were merged and termed `Other'. Lower and upper are the lower and upper bounds of the 99
\end{minipage}
\end{table}

Our findings indicated that the distribution of AI-generated data is
significantly distinct from that of the data in each individual
category, as well as from the dataset as a whole. In every instance, the
MMD statistic significantly exceeds the 99\% confidence interval,
underscoring the distinctiveness of the AI-generated titles.

\begin{figure}[htbp]
\centering
\includegraphics[width=0.8\linewidth]{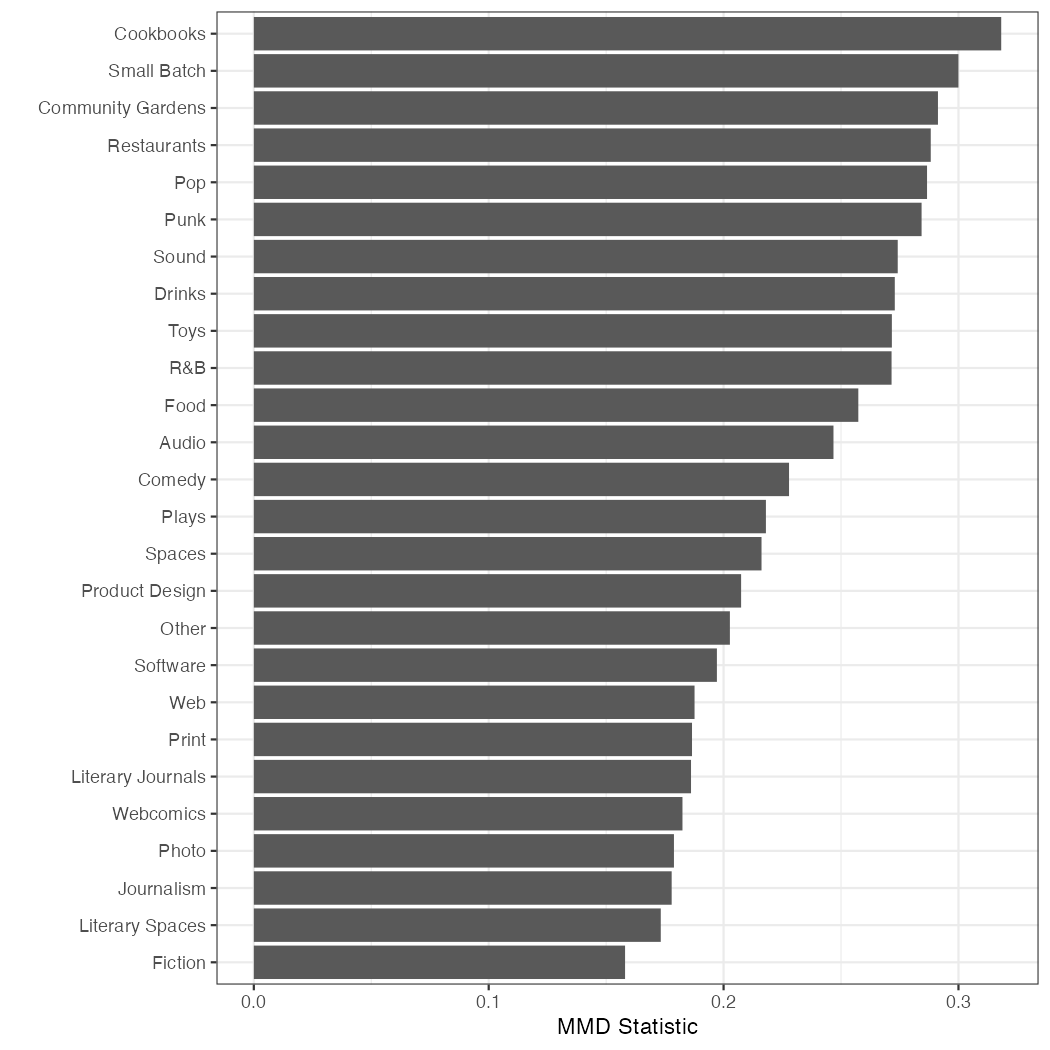}
\caption{Waterfall Plot of Category-Specific Effects}
\label{fig:waterfall_categories}
\begin{minipage}{\linewidth}
\medskip
\footnotesize
Note: The category-specific effects are estimates of the MMD statistic when applied to category specific field data and the AI-generated data.
\end{minipage}
\end{figure}

Additionally, we observed notable differences across categories. Figure
\ref{fig:waterfall_categories} illustrates these category-specific
effects, arranged in ascending order of their MMD estimates. This
arrangement places categories with the smallest effect
sizes---indicating greater similarity to the AI-generated data---on the
bottom. Conversely, categories with the largest effect sizes---denoting
lesser similarity to the AI-generated data---appear on the top.

The figure demonstrates that the categories most akin to the
AI-generated data include Fiction, Journalism, Literary Journals,
Literary Spaces, Photo, Print, Web, and Webcomics. This suggests that
the AI-generated titles closely align with media and written arts in
terms of their thematic focus. Conversely, the categories that diverged
most significantly from the AI-generated data encompass Audio, Community
Gardens, Cookbooks, Food, Pop, Punk, R\&B, Restaurants, Small Batch, and
Toys. These categories typically involve music, food, gardening, or
toys---themes prevalent in crowdfunding but less represented in the
AI-generated titles. Furthermore, the fantastical themes prevalent in
the AI-generated data seem to resonate more with fiction, while
categories like gardening are more grounded in practicality and
realistic imagery, a distinction clearly highlighted by the MMD metric.

These observations are particularly noteworthy given that the AI was not
directed to prioritize any specific categories. The AI's focus appears
to be inherently shaped by its training. This could be attributed to the
abundance of written material in these domains, providing a rich dataset
for a language model AI to learn from. On the other hand, categories
like music and food, which may depend more on visual stimuli, might not
have been as heavily emphasized during the AI's training.

Next, we investigated how the timing of AI data generation affects our
results, as presented in Table \ref{tab:findings_window}. The analysis
divides the data into 12 sequential windows, each containing 500 titles.
The table lists these windows in rows and provides the estimated MMD
statistic for each, along with its lower and upper confidence bounds.

\begin{table}[htbp]
\centering
\begin{tabular}{lrrr}
\toprule
& Estimate & Lower & Upper \\
\midrule
1-500 & 0.12 & > -0.01 & < 0.01 \\
501-1000 & 0.13 & > -0.01 & < 0.01 \\
1001-1500 & 0.13 & > -0.01 & < 0.01 \\
1501-2000 & 0.15 & > -0.01 & < 0.01 \\
2001-2500 & 0.15 & > -0.01 & < 0.01 \\
2501-3000 & 0.16 & > -0.01 & < 0.01 \\
3001-3500 & 0.16 & > -0.01 & < 0.01 \\
3501-4000 & 0.16 & > -0.01 & < 0.01 \\
4001-4500 & 0.16 & > -0.01 & < 0.01 \\
4501-5000 & 0.17 & > -0.01 & < 0.01 \\
5001-5500 & 0.17 & > -0.01 & < 0.01 \\
5501-6000 & 0.17 & > -0.01 & < 0.01 \\
\bottomrule
\end{tabular}
\caption{Window-Specific MMD Statistic}
\label{tab:findings_window}
\begin{minipage}{\linewidth}
\medskip
\footnotesize
Note: Estimate is the MMD statistic comparing the window-specific AI generated data to the observed field data. The window-specific measures are computed for 12 non-overlapping windows of 500 project titles each. Lower and upper are the lower and upper bounds of the 99
\end{minipage}
\end{table}

We found that the initial data generated by the AI closely resembled the
observed field data. However, as the AI was tasked with creating
increasingly distinct project titles---specifically, titles that
diverged from all previously generated titles---the divergence from the
field data became more pronounced, as evidenced by the rising MMD
statistic. This observation aligns with our findings presented in Figure
\ref{fig:entropy}, which show that as the AI was pushed to generate a
greater number of project titles, the entropy of the generated titles
increased. This suggests that the AI employed more infrequently used
words to present more esoteric ideas. As a result, the initial titles
were more similar to the field data.

\hypertarget{contributions-and-conclusion}{%
\section{Contributions and
Conclusion}\label{contributions-and-conclusion}}

In this paper, we make several important contributions to the literature
(\protect\hyperlink{ref-ding2020logical}{Ding 2020},
\protect\hyperlink{ref-ma2020machine}{Ma and Sun 2020}). First, we
address a significant gap in our understanding of AI's capacity for
novelty and creativity. While prior research has examined AI's potential
for automating various tasks
(\protect\hyperlink{ref-davenport2020artificial}{Davenport et al. 2020},
\protect\hyperlink{ref-huang2021strategic}{Huang and Rust 2021}), there
has been limited empirical investigation into AI's ability to generate
truly novel and unique content. By focusing on crowdfunding and using a
state-of-the-art language model, we provide compelling evidence that AI
can indeed generate ideas that are distinct from human-generated
examples and from each other, thereby challenging notions of AI as
merely a `stochastic parrot.' This finding not only demonstrates AI's
potential in enhancing and complementing human innovation, it
underscores its transformative potential in reshaping information
systems---offering novel solutions in the way information is generated,
processed, and utilized.

Second, we introduce a novel methodological approach to compare the
distributions of unstructured textual data. This approach employs KMEs,
which have primarily been used for structured data in natural language
data. The proposed method enables researchers to quantitatively assess
the similarity between samples drawn from different text distributions,
such as AI-generated and human-written content. This methodological
innovation has broad potential applications where unstructured data is
prevalent.

Third, we contribute to ongoing debates in the literature on the nature
of creativity and the role of AI in creative industries. Our results
suggest that AI's capacity for novelty increases as it is pushed to
generate more content, but also that there may be limits to this
capacity. The AI-generated titles were found to align more closely with
certain real-world categories (e.g., media and written arts) than others
(e.g., music and food), indicating that AI's creative output may be
influenced by biases in its training data. These nuances highlight the
need for further research on the factors that shape AI's creative
capabilities and how they might interact with human creativity.

For managers considering the use of AI, our findings present both
opportunities and challenges. On the one hand, they suggest that AI
could be a powerful tool for ideation and creative problem-solving,
potentially streamlining the creative process and leading to more
innovative campaigns. On the other hand, the question of legal
responsibility for AI-generated content remains a gray area. Even as AI
demonstrates a significant capacity for novelty, if an AI-generated idea
is found to infringe on existing copyrights or trademarks, it is unclear
where the liability would fall.

At a societal level, our findings contribute to the ongoing debate about
the role and status of AI-generated content. As AI's creative
capabilities continue to advance, issues of intellectual property
rights, attribution, and accountability are becoming increasingly
pressing. If AI-generated ideas are truly novel, it raises questions
about whether and how they should be protected under intellectual
property laws. There may be a need to develop new legal frameworks that
acknowledge the unique nature of AI creativity and provide appropriate
protections and attributions. This is a complex issue that requires
further discussion and collaboration among policymakers, legal experts,
and industry stakeholders.

\newpage

\hypertarget{web-appendix-a-examples-of-generated-project-titles}{%
\section{Web Appendix A: Examples of Generated Project
Titles}\label{web-appendix-a-examples-of-generated-project-titles}}

The main manuscript explores the capacity of artificial intelligence
(AI) to generate novel marketing communications, with a specific focus
on crowdfunding project titles. This Web Appendix supplements analysis
by providing 30 examples of AI-generated project titles. These titles
were produced by GPT-4-Turbo, a state-of-the-art language model, as part
of an experimental setup designed to gauge the AI's creativity.

\begin{table}[htbp]
\centering
\begin{tabularx}{0.66 \textwidth}{lX}
\toprule
\textbf{No.} & \textbf{Project Titles} \\
\midrule
1 & Dreams of the Sky: A Hot Air Balloon Adventure  \\
2 & Nebula Echoes: A Sci-fi Graphic Novel  \\
3 & Harmony's Light: A Handcrafted Lantern Festival  \\
4 & Pixel Kingdom: The Ultimate Retro Video Game \\
5 & Whispering Shadows: An Urban Fantasy Thriller \\
6 & Waves of Sound: The Next Generation Music App \\
7 & Guardians of the Green: An Eco-Friendly Board Game \\
8 & Magical Kingdoms: An Augmented Reality Puzzle Adventure \\
9 & Rustic Brews: A Craft Beer Making Kit \\
10 & Starlight Voyages: An Interactive Space Opera \\
11 & Origins of Olympus: A Mythological Strategy Game \\
12 & Echoes of the Past: A Historical Documentary Series \\
13 & Creatures of the Deep: An Underwater Exploration Game \\
14 & Vintage Vibes: A Classic Car Restoration Project \\
15 & Pathways of the Mind: A Psychological Thriller Novel \\
16 & Silver Screen Dreams: An Independent Film Project \\
17 & Spectrum: An Art Installation Celebrating Diversity \\
18 & Tales from the Cryptid: A Monster Hunting Adventure \\
19 & Infinite Imagination: A Children's Storybook App \\
20 & Chronicles of the Cursed: A Dark Fantasy Comic Series \\
21 & Mystical Forest: An Enchanted Board Game \\
22 & Northern Lights: A Photographic Journey \\
23 & Forgotten Realms: An Archaeological Adventure \\
24 & Pioneers of the Lost World: A Survival Video Game \\
25 & Electric Dreams: Building the Future of Energy \\
26 & Echoes in the Void: A Space Exploration Graphic Novel \\
27 & Ancient Whispers: A Virtual Reality Mystery \\
28 & Thunderstrike: A Superhero Role-Playing Game \\
29 & Artisan’s Alley: A Craftsmanship Documentary \\
30 & Dragon’s Breath: A Fantasy Battle Arena Game \\
\bottomrule
\end{tabularx}
\caption{Examples of AI-Generated Project Titles}
\label{table:comparison}
\begin{minipage}{\textwidth}
\medskip
\footnotesize
Note: The first 30 project titles generated by GPT-4-Turbo.
\end{minipage}
\end{table}

\newpage

\hypertarget{web-appendix-b-comparing-ai-generated-and-field-data}{%
\section{Web Appendix B: Comparing AI-Generated and Field
Data}\label{web-appendix-b-comparing-ai-generated-and-field-data}}

Our methodology leverages the emerging literature in machine learning on
embeddings. An embedding is a mapping from a set of objects (such as
words or project titles) to a mathematical space (most typically a
vector space) such that object properties (e.g., semantic meaning) are
preserved in notions of distance and angle. The literature has taken two
paths in its discussion of embeddings. One strand of literature defines
more mathematical notions of embeddings (e.g.,
\protect\hyperlink{ref-sriperumbudur2010hilbert}{Sriperumbudur et al.
2010}). Here, the emphasis is on establishing formal properties that may
be useful in downstream analyses based on embeddings. Another strand of
literature seeks to discover and use machine learning embedding
algorithms (e.g., \protect\hyperlink{ref-mikolov2013efficient}{Mikolov
et al. 2013}). Here, the emphasis is on the development of algorithms
and establishing the empirical properties of the embeddings they
construct.

We employ existing approaches in the following ways. Leveraging advances
in machine learning embedding algorithms, we map project titles (more
broadly, marketing content) to an intermediate space such that location
and distance correspond to semantic meaning. To best match the outputs
of the AI, we advocate the use of the embedding algorithm that
corresponds to the AI, as exposed through calls to its encoder. Thus,
for GPT-4-Turbo, we employ `text-embedding-3-large'.

We relate this embedding to a Kernel Mean Embedding (KME) of the
distribution of project titles. Here, we employ the concept of a
Reproducing Kernel Hilbert Space (RKHS) of functions. Below, we provide
a discussion tailored to our use case; additional details can be found
within the referenced books and articles, with an exhaustive
presentation in Berlinet and Thomas-Agnan
(\protect\hyperlink{ref-berlinet2011reproducing}{2011}). We also refer
the interested reader to Muandet et al.
(\protect\hyperlink{ref-muandet2017kernel}{2017}) who provide an
accessible and thorough discussion of KME.

\hypertarget{rkhs-and-kme.}{%
\paragraph{RKHS and KME.}\label{rkhs-and-kme.}}

Pertinent to its use for comparing distributions, the defining
characteristic of an RKHS is its structure as a Hilbert space---a
concept from functional analysis that generalizes Euclidean
space---comprising functions with certain properties. Specifically:

\begin{enumerate}
\def\labelenumi{\arabic{enumi}.}
\item
  \textbf{Functions as Elements}: The elements of an RKHS are functions
  that map elements from an input domain to the real numbers
  (\(\mathbb{R}\)). The RKHS is a collection of such functions that
  share the space's defining properties. While the input domain may be a
  subset of \(\mathbb{R}^n\), for data of the form that we investigate,
  an RKHS is likely to be most useful when the input domain comprises
  non-numerical elements; if the AI's outputs are strictly numerical, a
  well-established literature in marketing and econometrics provides
  means for density estimation and testing.
\item
  \textbf{Hilbert Space Structure}: A Hilbert space is a complete vector
  space equipped with an inner product. This structure allows for the
  definition of geometric concepts such as angles and lengths (norms) in
  the space. An RKHS is a particular type of Hilbert space where the
  vectors are functions, and thus, it inherits all the properties of a
  Hilbert space, including completeness, which ensures that limits of
  sequences of functions in the space also belong to the space.
\item
  \textbf{Reproducing Property}: What distinguishes an RKHS from other
  function spaces is the reproducing property. For every function \(f\)
  in the RKHS and every point \(x\) in the domain, the value of the
  function at that point, \(f(x)\), can be reproduced by the inner
  product of \(f\) with a kernel function \(k(\cdot, x)\) associated
  with the space: \(f(x) = \langle f, k(\cdot, x) \rangle\). This kernel
  function \(k(x, y)\) is a bivariate function whose arguments are
  points in the domain, and it essentially provides a way to ``probe''
  the function \(f\) at any point \(x\) through the inner product.
\item
  \textbf{Kernel Functions}: The existence of a kernel function \(k\)
  that defines the inner product between functions in the space is
  fundamental. The kernel function itself captures the geometry and
  topology of the function space and allows for the implicit
  representation of high-dimensional or even infinite-dimensional
  feature spaces, which is a cornerstone of kernel methods in machine
  learning.
\item
  \textbf{Kernel Mean Embedding (KME)}: KME leverages the machinery of
  RKHS to embed probability distributions into a Hilbert space. Unlike
  the RKHS itself, which is inherently a space of functions, a KME
  represents a probability distribution as a single point (or vector) in
  the RKHS. Thus, while RKHS deals with functions, KME is about
  embedding distributions into the space defined by these functions.
\end{enumerate}

Specifically, given a probability distribution \(P\) over a domain
\(X\), and a reproducing kernel \(k\) that induces an RKHS
\(\mathcal{H}\), the KME of \(P\) into \(\mathcal{H}\) is defined as the
expected value of the feature maps (associated with \(k\)) over \(P\).
Mathematically, the embedding \(\mu_P\) of \(P\) is given by:

\[
\mu_P = \mathbb{E}_{X \sim P}[k(X, \cdot)] = \int_X k(x, \cdot) dP(x)
\]

Here, \(k(X, \cdot)\) represents the feature map associated with the
kernel \(k\), mapping samples \(X\) from the domain to functions in
\(\mathcal{H}\). \(\int_X k(x, \cdot) dP(x)\) is a Bochner integral.

\hypertarget{mmd-in-structured-data}{%
\paragraph{MMD in Structured Data}\label{mmd-in-structured-data}}

We form an RKHS through a characteristic Mercer kernel to enable the
computation of distances between distributions as the distances between
the corresponding KME in the RKHS without computing the high-dimensional
features or taking expectations. This process can be viewed as analogous
to the use of the kernel trick in support vector machines (SVM) to
compute distances between points in a high-dimensional feature space
without computing high-dimensional features
(\protect\hyperlink{ref-steinwart2008support}{Steinwart and Christmann
2008}). Specifically, we can distinguish between samples from two
distributions through the computation of the following test statistic
(\protect\hyperlink{ref-gretton2012kernel}{Gretton et al. 2012}):

\[
\text{MMD}^2(P, Q) = \left\| \frac{1}{m} \sum_{i=1}^{m} \phi(x_i) - \frac{1}{n} \sum_{j=1}^{n} \phi(y_j) \right\|^2_{\mathcal{H}}.
\]

Here, \(\phi(\cdot) = k(x, \cdot)\) is the feature map that embeds the
data into the RKHS \(\mathcal{H}\), \(x_i\) and \(y_j\) are samples from
distributions \(P\) and \(Q\) respectively, \(m\) and \(n\) are the
number of samples from \(P\) and \(Q\) respectively. The double bars
\(\left\| \cdot \right\|_{\mathcal{H}}\) denote the norm in the RKHS.

\hypertarget{mmd-in-unstructured-data}{%
\paragraph{MMD in Unstructured Data}\label{mmd-in-unstructured-data}}

If our input data were numerical, then the model structure described
above would suffice. However, since our data is non-numerical, we
consider a compositional structure:

\[
\phi(\cdot) = \phi_2(\phi_1(x, \cdot)).
\]

Here, \(\phi_1\) is an injective map from the non-numerical data to an
intermediate Banach space (a complete normed vector space), and
\(\phi_2\) is a map such that its composition with \(\phi_1\) yields
\(\phi(\cdot)\) (i.e., \(\phi_2\) is implicitly defined through
\(\phi_1\) and \(\phi_2\)). This structure has the following
interpretation: \(\phi_1\) serves to express non-numerical data
numerically in a vector space with sufficient dimensionality to
distinguish between any two distinct elements (i.e., in our study,
ensuring each project title has a distinct textual representation),
while \(\phi_2\) ensures that \(\phi\) possesses sufficient richness so
that \(\mu_P\) accurately describes \(P\). This approach yields the
statistic:

\[
\text{MMD}^2(P, Q) = \left\| \frac{1}{m} \sum_{i=1}^{m} \phi_2(\phi_1(x_i)) - \frac{1}{n} \sum_{j=1}^{n} \phi_2(\phi_1(y_j)) \right\|^2_{\mathcal{H}}.
\]

Finally, an application of the kernel trick enables this statistic to be
computed without requiring the computation of the compositional feature
map. Specifically, an unbiased estimate of \(MMD^2\) is provided by the
equation:

\[
\widehat{MMD}^2 (P, Q) = \frac{1}{m(m-1)} \sum_{i=1}^{m} \sum_{\substack{j=1 \\ j \neq i}}^{m} k(x_i, x_j) + \frac{1}{n(n-1)} \sum_{i=1}^{n} \sum_{\substack{j=1 \\ j \neq i}}^{n} k(y_i, y_j) - \frac{2}{mn} \sum_{i=1}^{m} \sum_{j=1}^{n} k(x_i, y_j)
\]

Here, the components of this equation are interpreted as follows:

\begin{itemize}
\tightlist
\item
  \(k(\cdot, \cdot)\) is the kernel function used within the RKHS. In
  our reporting, \(\phi_1\) is a mapping established by the AI, and
  \(\phi_2\) corresponds to the second-order homogeneous polynomial
  kernel. As the support is discrete, these suffice to form a
  characteristic kernel.
\item
  \(x_i\) and \(x_j\) are samples drawn from distribution \(P\).
\item
  \(y_i\) and \(y_j\) are samples drawn from distribution \(Q\).
\item
  \(m\) and \(n\) are the sample sizes for samples drawn from
  distributions \(P\) and \(Q\), respectively.
\item
  The first term
  \(\frac{1}{m(m-1)} \sum_{i=1}^{m} \sum_{\substack{j=1 \\ j \neq i}}^{m} k(x_i, x_j)\)
  calculates the average of the kernel evaluations over all unique pairs
  of samples from \(P\).
\item
  The second term
  \(\frac{1}{n(n-1)} \sum_{i=1}^{n} \sum_{\substack{j=1 \\ j \neq i}}^{n} k(y_i, y_j)\)
  calculates the average of the kernel evaluations over all unique pairs
  of samples from \(Q\).
\item
  The third term
  \(-\frac{2}{mn} \sum_{i=1}^{m} \sum_{j=1}^{n} k(x_i, y_j)\) subtracts
  twice the average of the kernel evaluations between samples from \(P\)
  and samples from \(Q\).
\end{itemize}

\newpage

\hypertarget{web-appendix-c-simulations}{%
\section{Web Appendix C: Simulations}\label{web-appendix-c-simulations}}

In this Web Appendix, we outline the performance of our proposed
methodology designed for comparing sets of unstructured data, with a
focus on collections of verbal documents. The simulations aim to assess
the methodology's capability to differentiate between sets of verbal
data generated from distinct stochastic processes.

For these simulations, we engaged an AI with two specific prompts: one
to generate brand names with a humorous undertone and another for brand
names with a solemn tone. This dichotomy was chosen because it mirrors
common distinctions in real-world brand naming practices---such as the
playful name of a pizza parlor versus the gravitas of a funeral home's
name.

This approach of instructing the AI to produce brand names under
contrasting thematic conditions---humorous versus solemn---serves as a
direct method to explore a broader issue: how varying instructions lead
to differences in the stochastic processes underlying the generation of
brand names. This principle can be applied to other scenarios where
differences arise from distinct reasons.

From this exercise, we collected 100 humorous and 100 solemn brand names
generated by the AI. A selection of these names is presented below in
Table \ref{table:brandnames}, showcasing the AI's ability to generate a
diverse range of thematic expressions.

\begin{table}[htbp]
\centering
\begin{tabular}{l l l}
\toprule
\textbf{No.} & \textbf{Type} & \textbf{Brand Names} \\
\midrule
1  & Funny & Loaf \& Devotion \\
2  & Funny & Bread Pitt \\
3  & Funny & Sofa So Good \\
4  & Funny & Lord of the Fries \\
5  & Funny & Wok This Way \\
6  & Funny & The Codfather \\
7  & Funny & Planet of the Grapes \\
8  & Funny & Indiana Jeans \\
9  & Funny & Thai Tanic \\
10 & Funny & Pita Pan \\
11 & Funny & Brewed Awakening \\
12 & Funny & Tequila Mockingbird \\
13 & Funny & Eggscellent \\
14 & Funny & Jurassic Pork \\
15 & Funny & Bean Me Up \\
16 & Solemn & Serene Streams \\
17 & Solemn & Eternal Oak \\
18 & Solemn & Harmony Haven \\
19 & Solemn & Tranquil Pathways \\
20 & Solemn & Noble Quest \\
21 & Solemn & Zenith Peak \\
22 & Solemn & Evermore Estates \\
23 & Solemn & Pinnacle Trust \\
24 & Solemn & Sage Wisdom \\
25 & Solemn & Infinite Horizons \\
26 & Solemn & Legacy Builders \\
27 & Solemn & Unity Financial \\
28 & Solemn & Virtue Ventures \\
29 & Solemn & Paramount Partners \\
30 & Solemn & Guardian Gate \\
\bottomrule
\end{tabular}
\caption{Examples of AI-Generated Brand Names}
\label{table:brandnames}
\begin{minipage}{\textwidth}
\medskip
\footnotesize
Note: A list of the first 15 funny and the first 15 solemn brand names generated by GPT-4-Turbo.
\end{minipage}
\end{table}

Our presented methodology aims to establish a test statistic that
relates any two sets of verbal documents, such as these brand names.
Thus, for instance, when presented with the brand names in Table
\ref{table:brandnames}, one might ask whether the first 15 brand names
in the table are distinct from the last 15 brand names. How would one
statistically test for such a difference?

\begin{figure}[htbp]
\centering
\begin{subfigure}{.65\textwidth}
\centering
\includegraphics[width=\linewidth]{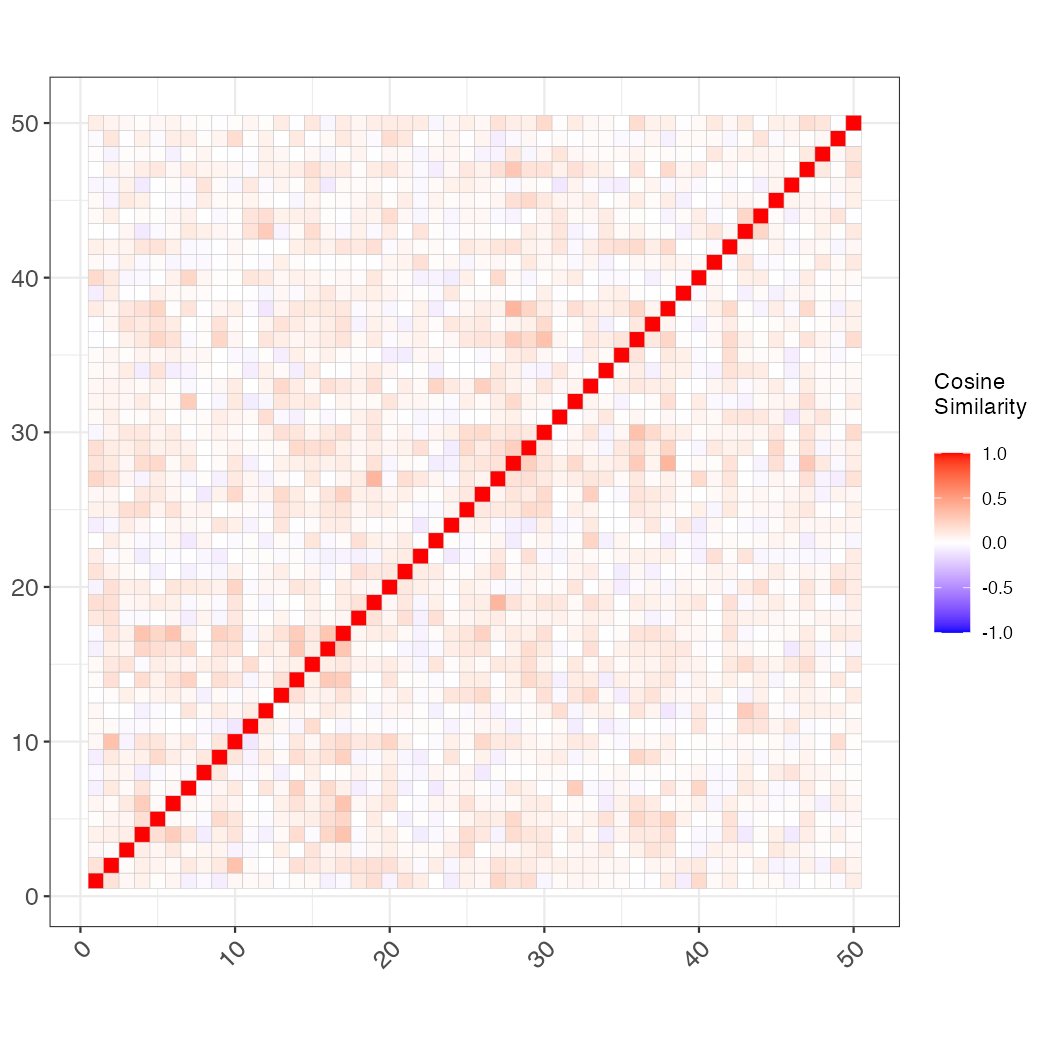}
\caption{Cosine Distance Among AI-Generated Funny Brand Names}
\label{fig:correlations1}
\end{subfigure}
\begin{subfigure}{.65\textwidth}
\centering
\includegraphics[width=\linewidth]{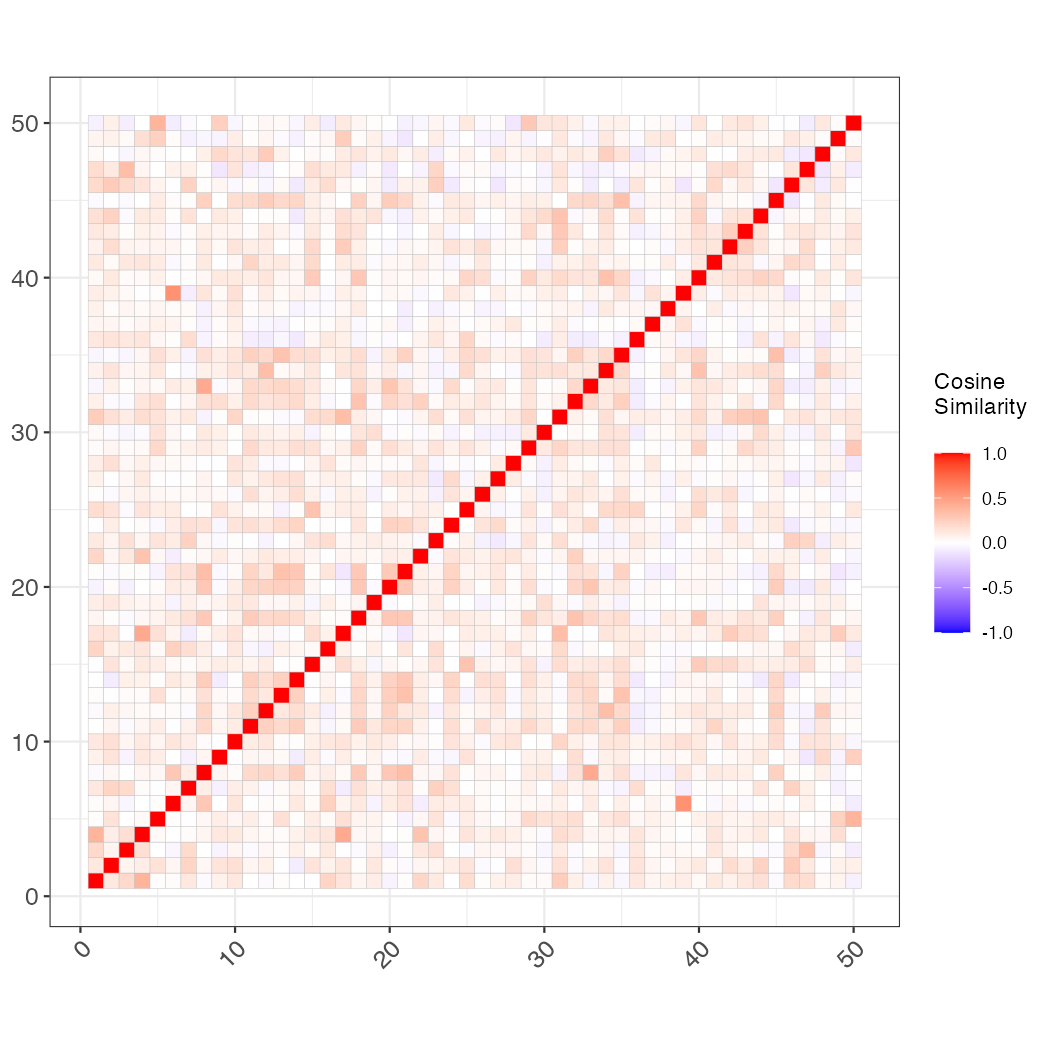}
\caption{Cosine Distance Among AI-Generated Solemn Brand Names}
\label{fig:correlations2}
\end{subfigure}
\caption{Cosine Distance Between AI-Generated Brand Names}
\label{fig:correlations}
\begin{minipage}{\linewidth}
\medskip
\footnotesize
Note: Cosine distance between embeddings of the brand names.
\end{minipage}
\end{figure}

To measure these differences, as a first step, we use the
`text-embedding-3-large' embedding, as also employed in our main
analysis. Figure \ref{fig:correlations} comprises two subfigures.
Subfigure \ref{fig:correlations1} visualizes the cosine distance of the
embedding of the first 50 funny brand names generated by the AI, where
we use only 50 names to make the figure easier to read. Subfigure
\ref{fig:correlations2} provides a similar visualization for the first
50 solemn brand names.

\begin{figure}[htbp]
\centering
\includegraphics[width=\linewidth]{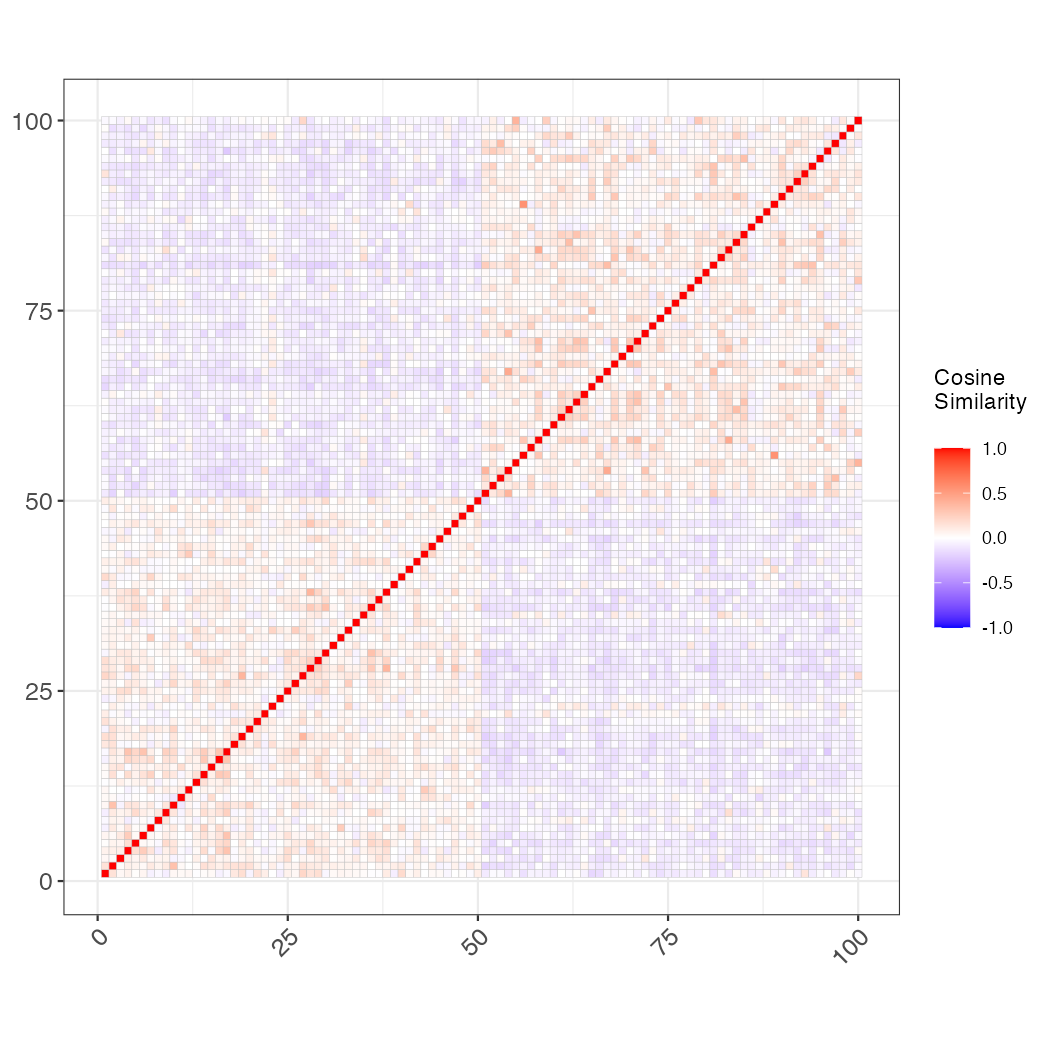}
\caption{Cosine Distance Between AI-Generated Funny and Solemn Brand Names}
\label{fig:correlations3}
\begin{minipage}{\linewidth}
\medskip
\footnotesize
Note: Cosine distance between embeddings of the brand names. The first 50 brand names are funny, and the last 50 are solemn.
\end{minipage}
\end{figure}

Figure \ref{fig:correlations3} visualizes the first 50 funny and 50
solemn brand names together. In all figures, the color of each square
indicates how similar or dissimilar each pair of brand names is, as
measured using cosine similarity, which is a typical measure of
similarity in machine learning.

We observe that the embeddings of the brand names show systematic
differences, whereby the funny brand names exhibit high cosine
similarity among themselves, and the solemn brand names exhibit high
cosine similarity among themselves. However, the two blocks of brand
names are highly dissimilar in terms of cosine dissimilarity. Thus, the
figure demonstrates that the use of the embedding enables the
identification of the pairwise similarity and dissimilarity of each
document in each collection---each brand name in each set of brand
names.

While there exist several alternatives for measuring pairwise distances,
there are few alternatives for comparing two sets of verbal documents.
This is because classic measures such as Euclidean and cosine distance
operate pairwise, and topic models seek to derive broader themes that
can be compared between sets of documents. In measuring the distance
between distributions, we aim to measure both the themes discussed in
the documents and the words used to discuss them. This is equivalent to
deriving a distance in both topic intensities and word-topic
distributions in topic models, while accounting for estimation
inaccuracies in the measurement of both distributions and without
imposing structure on the distance function employed on these
distributions.

In contrast, the MMD metric enables us to compare any set of brand names
to another set of brand names without knowledge of identity to establish
if the sets of brand names are distinct. When employed against the set
of funny and solemn brand names, this test translates to measuring if
the data-generating process through which the funny brand names were
generated is statistically distinct from the data-generating process
through which the solemn names were generated.

\begin{figure}[htbp]
\centering
\includegraphics[width=\linewidth]{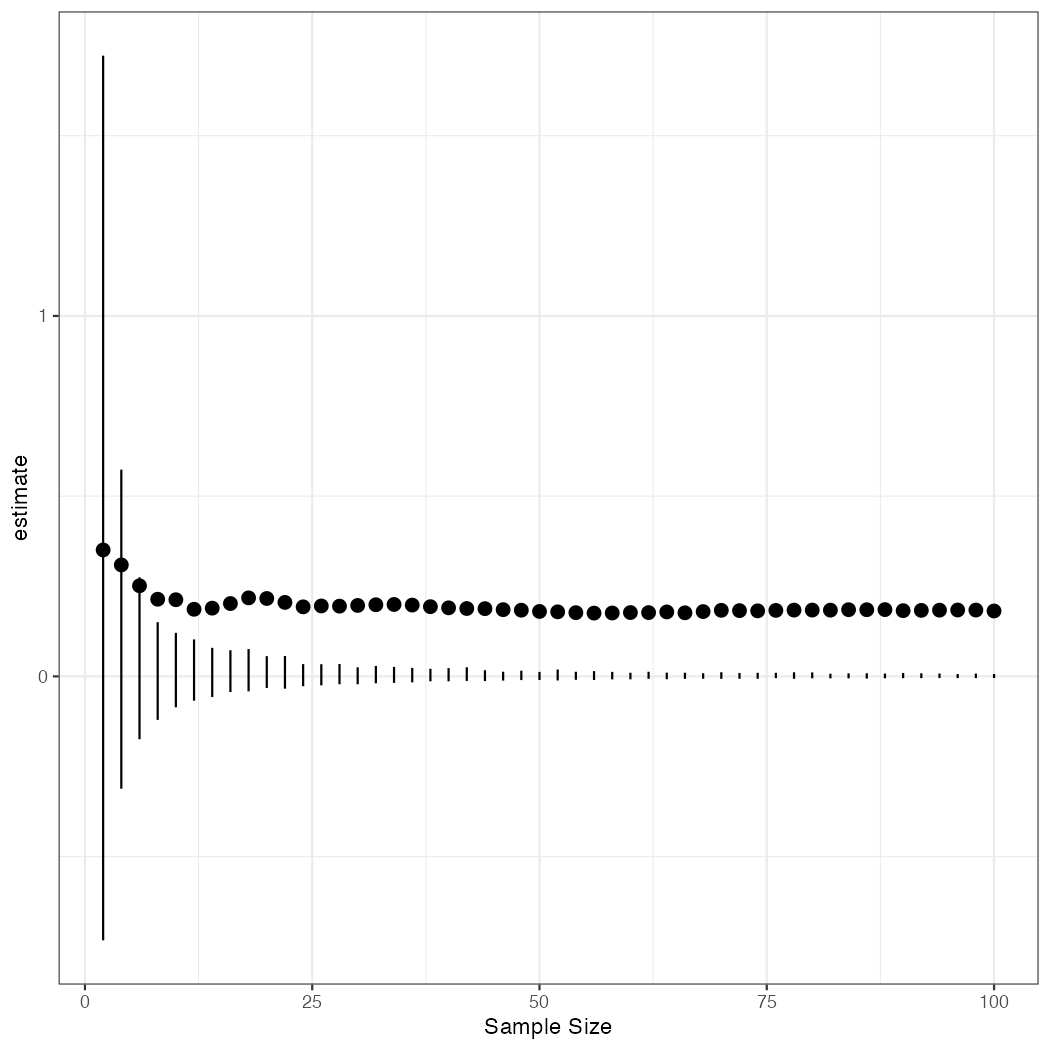}
\caption{MMD Statistic and Confidence Interval for Varying Sample Sizes}
\label{fig:simulation_results}
\begin{minipage}{\linewidth}
\medskip
\footnotesize
Note: MMD statistic and 99\% confidence interval for varying sample sizes. The sample size is on the x-axis. The statistic and confidence interval are represented on the y-axis. In cases where the point estimate falls outside the confidence interval, the data rejects the null hypothesis that both samples are from the same distribution.
\end{minipage}
\end{figure}

Note, the method does not require the set of samples to be
`comprehensive' in the sense of being the complete data---any 10
observations in our simulation from the set are sufficient to indicate
if the distributions are the same. This is because fundamentally the
method is designed for situations where samples are taken from a
distribution.

This is a crucial requirement of comparing any set of documents to the
output of generative AI, as the AI can always be used to generate more
data, and where our field data is representative but not
`complete'---there are other projects that are not covered in our data.
The use of classical methods such as pairwise cosine distance might have
been difficult to reconcile with the idea of the data being a
representative sample. However, as the distribution-based test provides
a comparison at the level of the data-generating process through the
likelihood of the generated words, it does not require either sample to
be complete for the method to provide substantive guidance.

\newpage
\singlespacing

\hypertarget{bibliography}{%
\section{Bibliography}\label{bibliography}}

\hypertarget{refs}{}
\begin{CSLReferences}{1}{0}
\leavevmode\vadjust pre{\hypertarget{ref-amabile2011componential}{}}%
Amabile T (2011) \emph{Componential theory of creativity} (Harvard
Business School Boston, MA).

\leavevmode\vadjust pre{\hypertarget{ref-amabile1988model}{}}%
Amabile TM et al. (1988) A model of creativity and innovation in
organizations. \emph{Research in organizational behavior}
10(1):123--167.

\leavevmode\vadjust pre{\hypertarget{ref-bender2021dangers}{}}%
Bender EM, Gebru T, McMillan-Major A, Shmitchell S (2021) On the dangers
of stochastic parrots: Can language models be too big? \emph{Proceedings
of the 2021 ACM conference on fairness, accountability, and
transparency}. 610--623.

\leavevmode\vadjust pre{\hypertarget{ref-berlinet2011reproducing}{}}%
Berlinet A, Thomas-Agnan C (2011) \emph{Reproducing kernel hilbert
spaces in probability and statistics} (Springer Science \& Business
Media).

\leavevmode\vadjust pre{\hypertarget{ref-boden2004creative}{}}%
Boden MA (2004) \emph{The creative mind: Myths and mechanisms}
(Psychology Press).

\leavevmode\vadjust pre{\hypertarget{ref-chomsky2023noam}{}}%
Chomsky N, Roberts I, Watumull J (2023) The false promise of ChatGPT.
\emph{The New York Times} 8.

\leavevmode\vadjust pre{\hypertarget{ref-Copyleaks_2024}{}}%
Copyleaks (2024)
\href{https://copyleaks.com/about-us/media/copyleaks-research-finds-nearly-60-of-gpt-3-5-outputs-contained-some-form-of-plagiarized-content}{Copyleaks
research finds nearly 60}. \emph{Copyleaks}.

\leavevmode\vadjust pre{\hypertarget{ref-davenport2020artificial}{}}%
Davenport T, Guha A, Grewal D, Bressgott T (2020) How artificial
intelligence will change the future of marketing. \emph{Journal of the
Academy of Marketing Science} 48:24--42.

\leavevmode\vadjust pre{\hypertarget{ref-ding2020logical}{}}%
Ding M (2020) \emph{Logical creative thinking methods} (Routledge).

\leavevmode\vadjust pre{\hypertarget{ref-felten2023occupational}{}}%
Felten EW, Raj M, Seamans R (2023) Occupational heterogeneity in
exposure to generative ai. \emph{Available at SSRN 4414065}.

\leavevmode\vadjust pre{\hypertarget{ref-fodor1988connectionism}{}}%
Fodor JA, Pylyshyn ZW (1988) Connectionism and cognitive architecture: A
critical analysis. \emph{Cognition} 28(1-2):3--71.

\leavevmode\vadjust pre{\hypertarget{ref-franceschelli2022copyright}{}}%
Franceschelli G, Musolesi M (2022) Copyright in generative deep
learning. \emph{Data \& Policy} 4:e17.

\leavevmode\vadjust pre{\hypertarget{ref-garon2023practical}{}}%
Garon J (2023) A practical introduction to generative AI, synthetic
media, and the messages found in the latest medium. \emph{Synthetic
Media, and the Messages Found in the Latest Medium (March 14, 2023)}.

\leavevmode\vadjust pre{\hypertarget{ref-girotra2023ideas}{}}%
Girotra K, Meincke L, Terwiesch C, Ulrich KT (2023) Ideas are dimes a
dozen: Large language models for idea generation in innovation.
\emph{Available at SSRN 4526071}.

\leavevmode\vadjust pre{\hypertarget{ref-gretton2012kernel}{}}%
Gretton A, Borgwardt KM, Rasch MJ, Schölkopf B, Smola A (2012) A kernel
two-sample test. \emph{The Journal of Machine Learning Research}
13(1):723--773.

\leavevmode\vadjust pre{\hypertarget{ref-harreis2023generative}{}}%
Harreis H, Koullias T, Roberts R, Te K (2023) Generative AI: Unlocking
the future of fashion. \emph{McKinsey \& Company}.

\leavevmode\vadjust pre{\hypertarget{ref-huang2021strategic}{}}%
Huang MH, Rust RT (2021) A strategic framework for artificial
intelligence in marketing. \emph{Journal of the Academy of Marketing
Science} 49:30--50.

\leavevmode\vadjust pre{\hypertarget{ref-huh2023chatgpt}{}}%
Huh J, Nelson MR, Russell CA (2023) ChatGPT, AI advertising, and
advertising research and education. \emph{Journal of Advertising}
52:477--482.

\leavevmode\vadjust pre{\hypertarget{ref-ivcevic2007artistic}{}}%
Ivcevic Z (2007) Artistic and everyday creativity: An act-frequency
approach. \emph{The Journal of Creative Behavior} 41(4):271--290.

\leavevmode\vadjust pre{\hypertarget{ref-lake2023human}{}}%
Lake BM, Baroni M (2023) Human-like systematic generalization through a
meta-learning neural network. \emph{Nature} 623(7985):115--121.

\leavevmode\vadjust pre{\hypertarget{ref-lemley2023generative}{}}%
Lemley MA (2023) How generative ai turns copyright law on its head.
\emph{Available at SSRN 4517702}.

\leavevmode\vadjust pre{\hypertarget{ref-ma2020machine}{}}%
Ma L, Sun B (2020) Machine learning and AI in marketing--connecting
computing power to human insights. \emph{International Journal of
Research in Marketing} 37(3):481--504.

\leavevmode\vadjust pre{\hypertarget{ref-mccoy2023much}{}}%
McCoy RT, Smolensky P, Linzen T, Gao J, Celikyilmaz A (2023) How much do
language models copy from their training data? Evaluating linguistic
novelty in text generation using raven. \emph{Transactions of the
Association for Computational Linguistics} 11:652--670.

\leavevmode\vadjust pre{\hypertarget{ref-mikolov2013efficient}{}}%
Mikolov T, Chen K, Corrado G, Dean J (2013) Efficient estimation of word
representations in vector space. \emph{arXiv preprint arXiv:1301.3781}.

\leavevmode\vadjust pre{\hypertarget{ref-muandet2017kernel}{}}%
Muandet K, Fukumizu K, Sriperumbudur B, Schölkopf B, et al. (2017)
Kernel mean embedding of distributions: A review and beyond.
\emph{Foundations and Trends{\textregistered} in Machine Learning}
10(1-2):1--141.

\leavevmode\vadjust pre{\hypertarget{ref-mumford2011handbook}{}}%
Mumford MD (2011) \emph{Handbook of organizational creativity} (Academic
Press).

\leavevmode\vadjust pre{\hypertarget{ref-Palmer_2023}{}}%
Palmer A (2023)
\href{https://www.cnbc.com/2023/03/28/amazon-sellers-are-using-chatgpt-to-help-write-product-listings.html}{Amazon
sellers are using chatgpt to help write product listings in sprawling
marketplace}. \emph{CNBC}.

\leavevmode\vadjust pre{\hypertarget{ref-pearl2018ai}{}}%
Pearl J, Mackenzie D (2018) AI can't reason why. \emph{Wall Street
Journal}.

\leavevmode\vadjust pre{\hypertarget{ref-samuelson2023generative}{}}%
Samuelson P (2023) Generative AI meets copyright. \emph{Science}
381(6654):158--161.

\leavevmode\vadjust pre{\hypertarget{ref-simonton2000creativity}{}}%
Simonton DK (2000) Creativity: Cognitive, personal, developmental, and
social aspects. \emph{American psychologist} 55(1):151.

\leavevmode\vadjust pre{\hypertarget{ref-sriperumbudur2010hilbert}{}}%
Sriperumbudur BK, Gretton A, Fukumizu K, Schölkopf B, Lanckriet GR
(2010) Hilbert space embeddings and metrics on probability measures.
\emph{The Journal of Machine Learning Research} 11:1517--1561.

\leavevmode\vadjust pre{\hypertarget{ref-steinwart2008support}{}}%
Steinwart I, Christmann A (2008) \emph{Support vector machines}
(Springer Science \& Business Media).

\leavevmode\vadjust pre{\hypertarget{ref-sternberg1999handbook}{}}%
Sternberg RJ (1999) \emph{Handbook of creativity} (Cambridge University
Press).

\leavevmode\vadjust pre{\hypertarget{ref-woodman1993toward}{}}%
Woodman RW, Sawyer JE, Griffin RW (1993) Toward a theory of
organizational creativity. \emph{Academy of management review}
18(2):293--321.

\leavevmode\vadjust pre{\hypertarget{ref-zhong2023copyright}{}}%
Zhong H, Chang J, Yang Z, Wu T, Mahawaga Arachchige PC, Pathmabandu C,
Xue M (2023) Copyright protection and accountability of generative ai:
Attack, watermarking and attribution. \emph{Companion proceedings of the
ACM web conference 2023}. 94--98.

\end{CSLReferences}

\end{document}